\documentclass{article}
\usepackage{microtype}
\usepackage{graphicx}
\usepackage{subfigure}
\usepackage{booktabs} 

\usepackage{hyperref}

\usepackage[accepted]{icml2023}

\usepackage{amsmath}
\usepackage{amssymb}
\usepackage{mathtools}
\usepackage{amsthm}

\usepackage[capitalize,noabbrev]{cleveref}

\theoremstyle{plain}

\theoremstyle{definition}

\theoremstyle{remark}

\renewcommand{\S} {\mathcal{S}}
\newcommand{\A} {\mathcal{A}}

\newcommand{\B} {\mathcal{B}}

\newcommand{\R} {\mathcal{R}}
\newcommand{\T}{\mathcal{T}}
\newcommand{\M} {\mathcal{M}}

\newcommand{\C}{\mathcal{C}}
\newcommand{\D}{\mathcal{D}}
\newcommand{\N}{\mathcal{N}}

\usepackage{float}

\DeclareMathOperator*{\argmax}{argmax} 
\DeclareMathOperator*{\argmin}{argmin} 

\usepackage[textsize=tiny]{todonotes}

\icmltitlerunning{Flipping Coins to Estimate Pseudocounts for
Exploration in Reinforcement Learning}

\begin{document}

\twocolumn[
\icmltitle{Flipping Coins to Estimate Pseudocounts\\for
Exploration in Reinforcement Learning}



\icmlsetsymbol{equal}{*}

\begin{icmlauthorlist}
\icmlauthor{Sam Lobel}{equal,yyy}
\icmlauthor{Akhil Bagaria}{equal,yyy}
\icmlauthor{George Konidaris}{yyy}
\end{icmlauthorlist}

\icmlaffiliation{yyy}{Department of Computer Science, Brown University, Providence, RI, USA}

\icmlcorrespondingauthor{Sam Lobel}{samuel\_lobel@brown.edu}
\icmlcorrespondingauthor{Akhil Bagaria}{akhil\_bagaria@brown.edu}

\icmlkeywords{Machine Learning, ICML}

\vskip 0.3in
]



\printAffiliationsAndNotice{\icmlEqualContribution} 

\begin{abstract}
We propose a new method for count-based exploration in high-dimensional state spaces. Unlike previous work which relies on density models, we show that counts can be derived by averaging samples from the Rademacher distribution (or \textit{coin flips}). This insight is used to set up a simple supervised learning objective which, when optimized, yields a state's visitation count. 
We show that our method is significantly more effective at deducing ground-truth visitation counts than previous work; when used as an exploration bonus for a model-free reinforcement learning algorithm, it outperforms existing approaches on most of $9$ challenging exploration tasks, including the Atari game \textsc{Montezuma's Revenge}.
\end{abstract}

\section{Introduction}
Deep exploration is crucial to solving long-horizon problems using reinforcement learning (RL) \citep{osband2016bootstrap}. When the number of states is small, an agent can simply keep track of how many times it has visited each state. This count can then be used as an exploration bonus to train a near-optimal policy \citep{strehl2008analysis}. When the world is much bigger than the agent, it may never revisit the same state \citep{sutton2022alberta}. To facilitate count-based exploration in such domains, the notion of visitation counts has been generalized to that of ``pseudocounts'' \citep{bellemare2016unifying} which behave similarly to counts but can be meaningfully applied in large or infinite state-spaces. Previous methods have equated the problem of estimating pseudocounts to the canonical machine learning problem of \textit{density estimation}: the more informative a given state is to the model while learning, the higher the reward for reaching it \citep{bellemare2016unifying, ostrovski2017count}.

While providing the first way to estimate pseudocounts, \citet{bellemare2016unifying}'s relationship between counts and probability densities exists only when the density model meets the following restrictions \citep{ostrovski2017count}:
\begin{itemize}
    \item It must output normalized probability densities, which precludes many powerful density models.
    \item It must be \textit{learning-positive}, which means that the probability density of a state must increase when it is encountered by the density model again.
    \item It must be updated exactly once per state visitation, precluding common techniques such as batching.
\end{itemize}

These requirements make density-based pseudocounts challenging to implement and sensitive to network architecture and hyperparameters such as learning rate. In light of these restrictions, it is tempting to forego count-based exploration in favor of other novelty estimates based on dynamics \citep{pathak2017curiosity} or observation \citep{Burda2018} prediction errors. But prediction-error based methods do not tell us how an exploration bonus should decay with repeated visits. Furthermore, they do not enjoy the same theoretical foundations afforded by count-based methods \citep{strehl2008analysis,mohammad2017minimax,jin2018q}.
For instance, count-based bonuses lead to near-optimal policies even when environments are highly stochastic; no such guarantees exist for prediction-error based methods.

We hypothesize that count-based exploration can be more effective than prediction-error based methods if we can compute pseudocounts under a less restrictive setting. Our core insight is that a state's visitation count can be derived from the sampling distribution of Rademacher trials made every time a state is encountered. We train a neural network, the \textit{Coin Flip Network} (CFN), to predict the average of this sampling distribution; by solving this supervised learning problem, we output the inverse of the state's visitation count. Unlike other pseudocount methods \citep{ostrovski2017count}, we do not place any restrictions on the type of function approximator or the procedure used to train it, thereby allowing a practitioner to select the model architecture best suited to their input modality.

We show that in visual versions of Gridworld \citep{allen2021learning} and \textsc{Taxi} \citep{dietterich1998maxq}, our method can recover the ground-truth counts while other pseudocount methods cannot. We then evaluate our algorithm on a variety of challenging sparse-reward continuous control problems; in these environments, we outperform baseline actor-critic \citep{haarnoja2018soft} and random network distillation \citep{Burda2018}, with the largest gains on the most challenging exploration domains. On the image-based exploration benchmark problem \textsc{Montezuma's Revenge} \citep{bellemare2013arcade}, we outperform baseline Rainbow \citep{hessel2018rainbow} and are competitive with state-of-the-art exploration algorithms \citep{ostrovski2017count,Burda2018}, which arguably have been overfit to this domain \citep{Taiga2020On}.
Finally, we show that increasing transition noise in Gridworld and \textsc{Montezuma's Revenge} causes RND's performance to degrade more rapidly than CFN's, as predicted by the theoretical properties of count-based exploration.

\section{Background and Related Work}

We consider problems modeled as Markov Decision Processes (MDPs) $\M=(\S,\A,\R,\T,\gamma)$ where $\S$ is the state-space, $\A$ is the action-space, $\R$ is a reward function, $\T$ is a transition function and $\gamma$ is the discount factor. The aim of the agent is to learn a policy that maximizes the expected discounted sum of rewards \citep{sutton2018reinforcement}.

\subsection{Deep reinforcement learning}
Model-free RL algorithms often use variants of Q-learning \citep{watkins1992q} to learn an action-value function $Q_{\theta}(s,a)$ and then act greedily with respect to it. This Q-function can be learned in high-dimensional spaces using non-linear
function approximators (parameterized by $\theta$) by minimizing the loss ${L(\theta)=\mathbb{E}[(Q_{\theta}(s,a)-y_t)^2]}$, where the Q-learning target $y_t$ is given by the following equation \citep{mnih2015human}:
$$y_t=\R_t+\gamma\max_{a_{t+1}}Q_{\theta'}(s_{t+1},a_{t+1}),$$
and $\theta'$ are the parameters of a slowly changing target network \citep{mnih2015human}. This principle has since been extended with various algorithmic improvements; for example, Rainbow \citep{hessel2018rainbow} for discrete action-spaces and Soft Actor Critic (SAC) \citep{haarnoja2018soft} for continuous action-spaces. The majority RL systems use myopic strategies for exploration (e.g, $\epsilon$-greedy, action noise), which do not scale to long-horizon problems. 

\subsection{Bonus-based exploration}
A promising strategy for exploration is to incorporate an \textit{intrinsic reward} that encourages the agent to gather informative data. 
This intrinsic reward $\B(s_t,a_t)$ is added to the extrinsic reward to create an augmented Q-learning target:
\begin{equation}
y_t = \R_t + \lambda\B(s_t,a_t) + \gamma \max_{a'}Q_{\theta'}(s_{t+1},a'),
\label{eq:bbe-q-target}
\end{equation}

where $\lambda \in \mathbb{R}^+$ modulates the scale of the exploration bonus. Acting greedily with respect to this Q-function balances exploration and exploitation, and is the basis for many provably efficient \cite{strehl2008analysis,jin2018q} and practically successful \cite{Taiga2020On} algorithms. As is typical in the literature, we consider bonuses $\B(s)$ that are only dependent on state \citep{burda2018large}.

\paragraph{Count-based exploration.} In tabular domains, a count-based exploration bonus of $1/\sqrt{N(s)}$ (near) optimally trades-off exploration and exploitation, even in highly stochastic MDPs \citep{auer2002using,strehl2008analysis}.
This approach was extended to function approximation by using density models to calculate pseudocounts: first with the CTS model \citep{bellemare2016unifying} and then with PixelCNN \citep{ostrovski2017count}. However, it is challenging to learn density models in high-dimensional observation spaces---especially given the restrictions discussed in the Introduction. Successor Counts \citep{machado2020count} also bypasses density modeling (by relating pseudocounts to the norm of successor representations \citep{dayan1993improving}). But in problems that require deep exploration \citep{Taiga2020On}, they are outperformed by PixelCNN \citep{machado2020count}, which we compare our method to in Section~\ref{sec:experiments}.

Many methods bypass learning and resort to workarounds that heavily incorporate domain knowledge---for example, $\#$-Counts \citep{tang2017exploration} and OPIQ \citep{Rashid2020Optimistic} use locality sensitive hashing, Go-explore \citep{ecoffet2021first} severely downsamples input images before binning them, and MEGA assumes knowledge of which dimensions of state are useful for the task \citep{pitis2020maximum}. By contrast, CFN takes raw observations as input and flexibly learns representations optimized for predicting exploration bonuses.

\paragraph{Model-prediction error.} Many methods learn a transition model and use the error in the predicted next state as an exploration bonus \citep{stadie2015incentivizing,houthooft2016vime,pathak2017curiosity,ermolov2020latent,guo2022byol}. This causes the agent to collect more data where the transition model is least accurate \citep{kearns2002near,brafman2002r,kakade2003exploration}.
However, most methods learn deterministic models \citep{pathak2017curiosity,guo2022byol}, making them difficult to scale to stochastic environments.

\paragraph{Random Network Distillation (RND).} Due to its simplicity and empirical strength, RND \citep{Burda2018} has emerged as the most popular exploration algorithm. RND assigns an exploration bonus to a state using a simple, elegant heuristic: the novelty of a state is directly proportional to how the accurately a trainable network can mimic a randomly-initialized network's projection of it. Compared to count-based methods, RND's exploration bonus is difficult to interpret---it is an unnormalized distance in a neural network latent space. Furthermore, it is unclear at what rate the RND bonus falls with visitation count: this depends on the learning dynamics of neural networks and stochastic gradient descent. In Section~\ref{sec:experiments} we investigate the shape of RND's bonus and compare it to CFN.

\subsection{Uncertainty Estimation} Some methods explicitly estimate epistemic uncertainty in the value function and use that to drive exploration \citep{osband2016bootstrap,osband2018randomized,o2018uncertainty}. These are promising realizations of Thompson sampling \citep{thompson1933likelihood} in high-dimensional domains, but empirically under-perform optimism driven approaches.

\subsection{Integrated exploratory RL agents} Novelty estimates are often integrated into larger RL agents that use more sophisticated techniques such as episodic memory \cite{badia2020never}, adaptive horizons \citep{badia2020agent57,kapturowski2022human}, transferable representation learning \citep{zhang2021noveld,raileanu2020ride}, goal-conditioned policies \citep{pong2019skew} or planning \citep{bagaria2021skill}. Our method improves how the novelty bonus itself is computed and can be included in any of these agents without much modification.

\section{Coin Flip Network (CFN)}\label{sec:method}

A pseudocount $\mathcal{N}: \S \rightarrow \mathbb{R}^+$ generalizes the notion of counts to large state-spaces and can be used to quantify the novelty of a state \citep{bellemare2016unifying}. Specifically, visiting a state $s$ affords the agent a novelty bonus of
${\B(s)=\frac{1}{\sqrt{\mathcal{N}(s)}}}$;
we will use a neural network, the \textit{Coin Flip Network} (CFN) $f_{\phi}$, to directly predict this count-based exploration bonus.

To learn $f_\phi$, we set up a simple regression problem:
\begin{equation}\label{eq:regression}
    f_\phi=\argmin_{\phi}\mathbb{E}_{(s_i,y_i)\sim\mathcal{D}}\Big[\mathcal{L}(s_i, y_i)\Big],    
\end{equation}
where $\mathcal{L}$ is the mean-square error loss function and $\mathcal{D}$ is a dataset of state-label pairs. Our main insights relate to the design of the labels $y_i$ in such a way that the resulting function $f_\phi$ will map each state to its count-based exploration bonus $\frac{1}{\sqrt{\mathcal{N}(s)}}$. 

\begin{figure}
    \centering
    \includegraphics[width=0.7\linewidth]{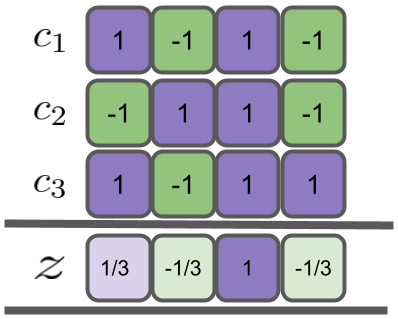}
    \vspace{0.2cm}
    \caption{Illustration of our counting method for a state $s$ with true count $3$. Each occurrence of $s$ creates new coin-flip vectors $c_1, c_2, c_3$. We average these vectors into $z$ and compute the squared magnitude. Dividing this by the number of coin flips $d=4$ yields the inverse count ${1/\N(s)}$.}
    \label{fig:cfn_block_illustration}
\end{figure}

\subsection{Counts from the Rademacher distribution}\label{sec:coin-flip_counts}

The first step in designing the labels $y_i$ in Eq~\ref{eq:regression} is to notice that moments of the Rademacher/coin-flip distribution directly encode counts.

Consider the fair coin-flip distribution $\C$ over outcomes $\{-1,1\}$. Imagine flipping this coin $n$ times, and averaging the results into $z_n$. In expectation, the average is $0$, but any given trial likely results in a non-zero value for $z_n$. Generally, for all $n$, the second moment of $z_n$ is related to the inverse-count:
\begin{equation}\label{eq:var-of-y}
    \mathcal{M}_2(z_n) = \mathbb{E}[z_n^2] = \sum_i Pr(z_n = i) * i^2 = 1/n.
\end{equation} 
This property is a simple restatement of the fact that $\mathbb{E}[z_n^2]$ is the \textit{variance of the sample mean} of the coin-flip distribution, which is well-known to scale inversely with sample size. In fact, this scaling is shared between all zero-mean unit-variance distributions, not just the coin-flip distribution. However, using this distribution leads to the \textit{lowest variance estimates of inverse-counts} out of the entire class of matching distributions. We prove these two facts in Appendix~\ref{sec:appendix-any-whitened-distribution-counts} and \ref{sec:appendix-variance-of-inverse-count-estimator} respectively.

\subsection{Estimating counts for a state via multiple coin flips}
An additional way to lower the variance of this estimator is to average together multiple estimates of $z^2_n$: by flipping $d$ coins each time, we get $d$ independent estimates of $\frac{1}{n}$, which reduces variance by a factor of $\frac{1}{d}$ (see Appendix~\ref{sec:appendix-more-flips-is-good}).

Consider the contrived case of an MDP with a single state and imagine that we draw a random sample from $\C^d$ each time that state is visited.
Equation~\ref{eq:var-of-y} implies that the squared magnitude of the averaged vectors is an unbiased estimator of $d/n$; this is also illustrated in Figure~\ref{fig:cfn_block_illustration}.
Of course, we do not need a novel method to count the number of elements in a list; but this informs our eventual method for producing bonuses in general MDPs.

\subsection{Predicting counts for multiple states}\label{subsec:count-for-multiple}
Having solved the uninteresting problem of extracting counts for a single state MDP, we will now generalize to datasets with multiple occurrences of multiple states. As label $y_i$ for state $s_i$ in Eq~\ref{eq:regression}, we generate a $d$-dimensional random vector $\mathbf{c_i}\sim\{-1,1\}^d$; this leads to the following simplification of Equation~\ref{eq:regression}:
\begin{align}\label{eq:defn-f-star}
f_{\phi}^*(s) &= \argmin_\phi \sum_{i=1}^{|\D|} \lVert \mathbf{c_i} - f_\phi(s_i)\rVert^2 \nonumber \\
       &= \argmin_\phi \sum_{i=1}^{|\D|} \sum_{j=1}^d (c_{ij} - f_\phi(s_i)_j)^2.
\end{align}

When there are multiple instances of the same state $s$ in $\D$, each occurrence will be paired with a different random vector. In that case, $f^*_\phi$ cannot learn a perfect mapping from states to labels, and instead minimizes $\mathcal{L}$ by outputting the \textit{mean} random vector for all instances of a given state:
\begin{align*}
    f_\phi^*(s) &= \frac{1}{n}\sum_{i=1}^n \mathbf{c_i}.
\end{align*}

Combining Equation~\ref{eq:var-of-y} and \ref{eq:defn-f-star} relates the solution $f_\phi^*$ to the inverse count:
\begin{align*}
    f_\phi^*(s) &= \frac{1}{n}\sum_{i=1}^n \mathbf{c_i}\\
    \implies \mathbb{E}\Big[\frac{1}{d}\lVert f_\phi^*(s)\rVert^2\Big] &= \frac{1}{d} \sum_{j=1}^d\mathbb{E}\Big[\Big(\sum_{i=1}^n\frac{c_{ij}}{n}\Big)^2\Big]&&\comment{Taking expectation of the vector norm}\\
    &= \frac{1}{d} \sum_{j=1}^d\mathbb{E}\Big[z_n^2\Big] &&\comment{Using Eq~\ref{eq:c-to-y}}\\
    &= \frac{1}{d} \sum_{j=1}^d\frac{1}{n} = \frac{1}{n}. &&\comment{Using Eq~\ref{eq:var-of-y}}
\end{align*}


Thus, by training $f_\phi$ on the objective described in Equation~\ref{eq:defn-f-star} we can map states to approximate count-based bonuses:

\begin{equation}
\boxed{
    \mathcal{B}(s) := \sqrt{\frac{1}{d}\lVert f_\phi(s)\rVert^2} \approx \frac{1}{\sqrt{\mathcal{N}(s)}}
}
\label{eq:bonus-fstar-count}
\end{equation}

\subsection{Generalizing outside the training data} \label{sec:linear}
%
The optimization procedure described so far will eventually derive the correct visitation counts for states in the training data (given a powerful function approximator and sufficient training iterations). But as the agent interacts with the environment, how will the CFN bonus generalize to states absent from the training data?
Although in practice we represent $f_\phi$ as a neural network, to gain intuition on generalization, we mathematically examine the case when $f_\phi$ is linear.
In this case, the bonus for a state $s$ is a linear combination
of the bonuses assigned to the right singular vectors of the training data; more discussion and proof is in Appendix~\ref{sec:linear-analysis}.
Intuitively, the singular vectors of the training data take on the role of unique states: instead of tracking how many of each state visitation there are, a linear $f_\phi$ records how much of each singular vector is present in total in the dataset.
Interestingly, when states are represented using one-hot vectors, the resulting solution to Equation~\ref{eq:defn-f-star} recovers tabular counts.

\subsection{Improving predictions for novel states}
As stated above, a learning architecture with infinite capacity would learn the exact inverse count for each unique state in the training data. But finite capacity and training time imply that the network will not learn this mapping exactly. Next, we will propose two ways to guide our network to favorably trade-off prediction errors among states in the dataset: first, we will use \textit{prioritized sampling} to preferentially learn the novelty of rare states; second, we will use \textit{optimistic initialization} to assign a pseudocount of $1$ to novel states newly added to the replay buffer. 

\subsubsection{Prioritizing Novel States}\label{sec:cfn_prioritization}

Since training to convergence at every time step is not feasible, we update $f_\phi$ once every time step on a mini-batch of states drawn from a replay buffer.
Revisiting the optimization problem from Equation~\ref{eq:defn-f-star}, we note that a state $s$ with count $n$ will appear in uniform sampling $n$ times more often than a state visited only once. This would make $f_\phi$ focus too much on learning the bonus of high count states, which are uninteresting from an exploration perspective.
To remedy this problem, we would like to assign more weight to low count states by sampling them with greater probability \cite{schaul2015prioritized}.

Of course, we do not have access to the true count during training; so, we approximate this procedure by prioritizing by our current \textit{estimate} of inverse-count:

$$\text{priority}(s) \leftarrow \frac{1}{d}\lVert f_\phi(s)\rVert^2\approx \frac{1}{\mathcal{N}(s)}.$$

Though this prioritization changes the importance of different states relative to each other, all instances of the same state will be sampled in equal proportion. Therefore, solving the prioritized version of Equation~\ref{eq:defn-f-star} still outputs the unbiased average of a state's coin-flip vectors (and therefore the correct pseudocounts). 

Prioritizing in this way introduces another difficulty: if a state has been recently added to the replay buffer, it has not appeared in many gradient updates and thus we cannot trust our estimate of its count. To combat this, we also prioritize sampling by the number of times, $n_{\text{updates}}(s)$, we have sampled $s$ in the past. We combine both these prioritization schemes using an $\alpha$-weighted sum (we use $\alpha=0.5$):
%

\begin{equation}\label{eq:priority}
  \text{priority}(s) = \alpha\left(\frac{1}{n_\text{updates}(s)}\right) + (1 - \alpha)\frac{1}{d}\lVert f_\phi(s)\rVert^2.  
\end{equation}


The $n_{\text{updates}}$ term  weighs different instances of the same state differently, but its effect on prioritization disappears quickly during training,  so it does not influence the fixed point either.

\subsubsection{Optimistic Initialization of Bonus}\label{sec:cfn_randomprior}

Consider a state $s$ that CFN has not been trained on yet. The exploration bonus $\B(s)$ will be determined by CFN's generalization properties. If $s$ is very different than the other states that CFN has already been trained on, then $\B(s)$ tends to be close to $0$, when in fact we would like the pseudocount of novel states to be initialized to $1$.

We achieve this optimistic initialization  using a \textit{random prior} \citep{osband2018randomized}:
$$f_{\phi}(s) = \hat{f}_\phi(s)+f_{\text{prior}}(s),$$
where $f_{\text{prior}}$ is the output of a frozen and randomly initialized neural network, and $\hat{f}_\phi$ is the trainable component of CFN.
We use a running mean and variance to normalize the prior so that $\mathbb{E}_{s\sim\mathcal{D}}[f^{(i)}_{\text{prior}}(s)^2] = 1$ over all output dimensions $i\in\{1,..,d\}$. This ensures that if $\hat{f}_\phi(s) = 0$ on a novel state $s$, then $\lVert f_{\phi}(s)\rVert^2 = 1$, i.e., states are added to the buffer with an approximate initial pseudocount of $1$. As training progresses, the effect of the initialization will wash out and $\B(s)$ will eventually settle to its correct value.  An analysis of the optimistic prior's contribution is  in Appendix~\ref{sec:app-zero-flips}.

\subsection{Integrated CFN agent}\label{sec:full_cfn_agent}
Algorithm~\ref{alg:cfn} outlines how CFN is combined with Rainbow \citep{hessel2018rainbow} to form a complete bonus-based exploration agent. Naturally, we can combine CFN with most off-the-shelf RL algorithms with minor changes \citep{haarnoja2018soft,mnih2015human,lillicrap2015continuous}.

\begin{algorithm}[t]
    \caption{Rainbow-CFN Agent}
    \label{alg:cfn}
    \textbf{Hyperparameters}: Reward scale $\lambda$, Number of coin flips $d$\\
    \begin{algorithmic}[1] 
        \STATE Initialize Q-network $Q_\theta$ and target Q-network $Q_{\theta'}$.
        \STATE Initialize CFN prior $f_{\text{prior}}$ and trainable network $\hat{f}_{\phi}$.
        \STATE Initialize replay buffer for Rainbow $B_{r}$ and CFN $B_c$.
        \STATE Initialize optimizer for Rainbow and for CFN.
        \STATE Initialize the running mean $\mu_t$ and variance $\sigma^2_t$ for the optimistic prior $f_{\text{prior}}$.
        \WHILE{training}
            \STATE $s_0 =$ \texttt{env.reset()}
            \WHILE {not \texttt{done}}
                \STATE $a_t = \argmax_{a\in\A}Q_\theta(s_t, a)$
                \STATE $R_t, s_{t+1},$ \texttt{done} = \texttt{env.step}($s_t,a_t$)
                \STATE Compute intrinsic reward $\B(s_t)$ using Equation~\ref{eq:bonus-fstar-count}.
                \STATE Update $\mu_t$ and $\sigma^2_t$ using $f_{\text{prior}}(s_t)$.
                \STATE Add transition $(s_t, a_t, R_t, \B(s_t), s_{t+1})$ to $B_r$.
                \STATE Sample a random coin-flip vector $\mathbf{c}\sim\{-1,1\}^d$.
                \STATE Add state coin-flip tuple $(s_t, \mathbf{c})$ to $B_c$.
                \STATE Sample minibatch $(s,a,r,\B_s,s')\sim B_r$ and update $Q_\theta$ using Rainbow's optimizer and Eq~\ref{eq:bbe-q-target}.
                \STATE Update priority for minibatch using Rainbow.
                \STATE Sample minibatch $(s,\mathbf{c}) \sim B_c$ and use CFN's optimizer to update $f_\phi$ via one gradient step on the loss function corresponding to Equation~\ref{eq:defn-f-star}.
                \STATE Update priority for minibatch using Equation~\ref{eq:priority}.
            \ENDWHILE
        \ENDWHILE
    \end{algorithmic}
\end{algorithm}

\section{Experiments}\label{sec:experiments}
\begin{figure*}
    \centering
    \includegraphics[width=\linewidth]{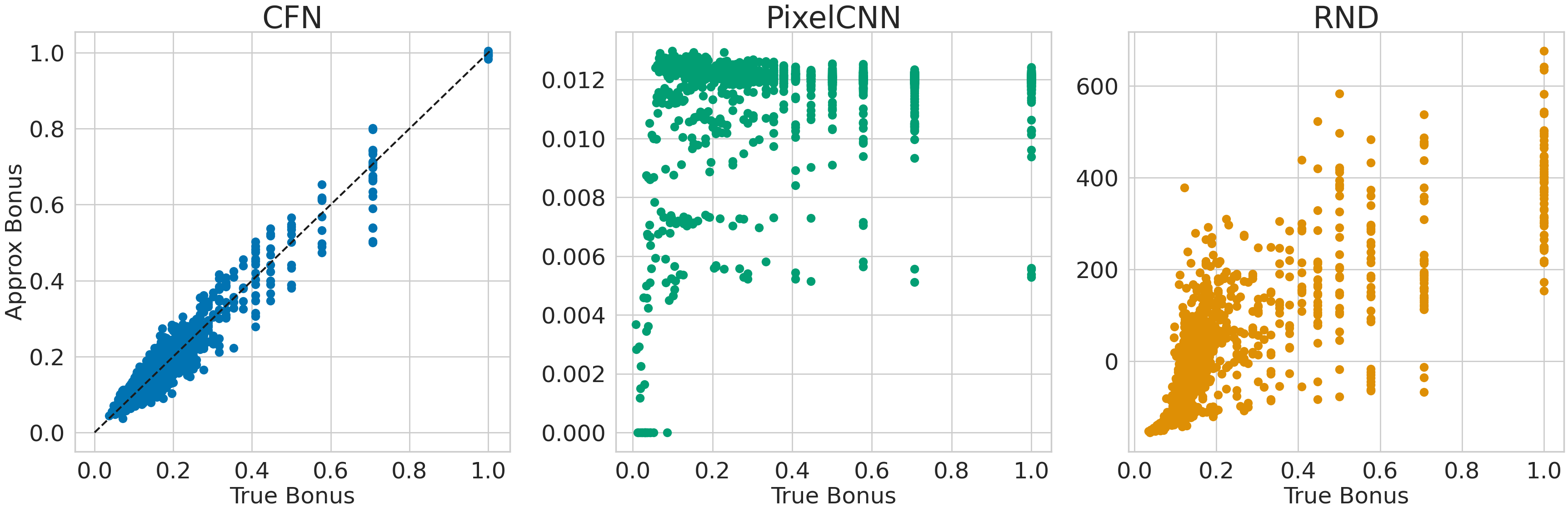}
    \vspace{-0.5cm}
    \caption{Predicted count-based bonuses for all three methods in Visual Gridworld after $100,000$ interactions. The horizontal axis is the ground truth $1/\sqrt{\mathcal{N}(s)}$ bonus, the vertical axis is the exploration bonus predicted by the different methods.}
    \label{fig:gridworld-counts}
\end{figure*}

\begin{figure}
    \centering
    \includegraphics[width=\linewidth]{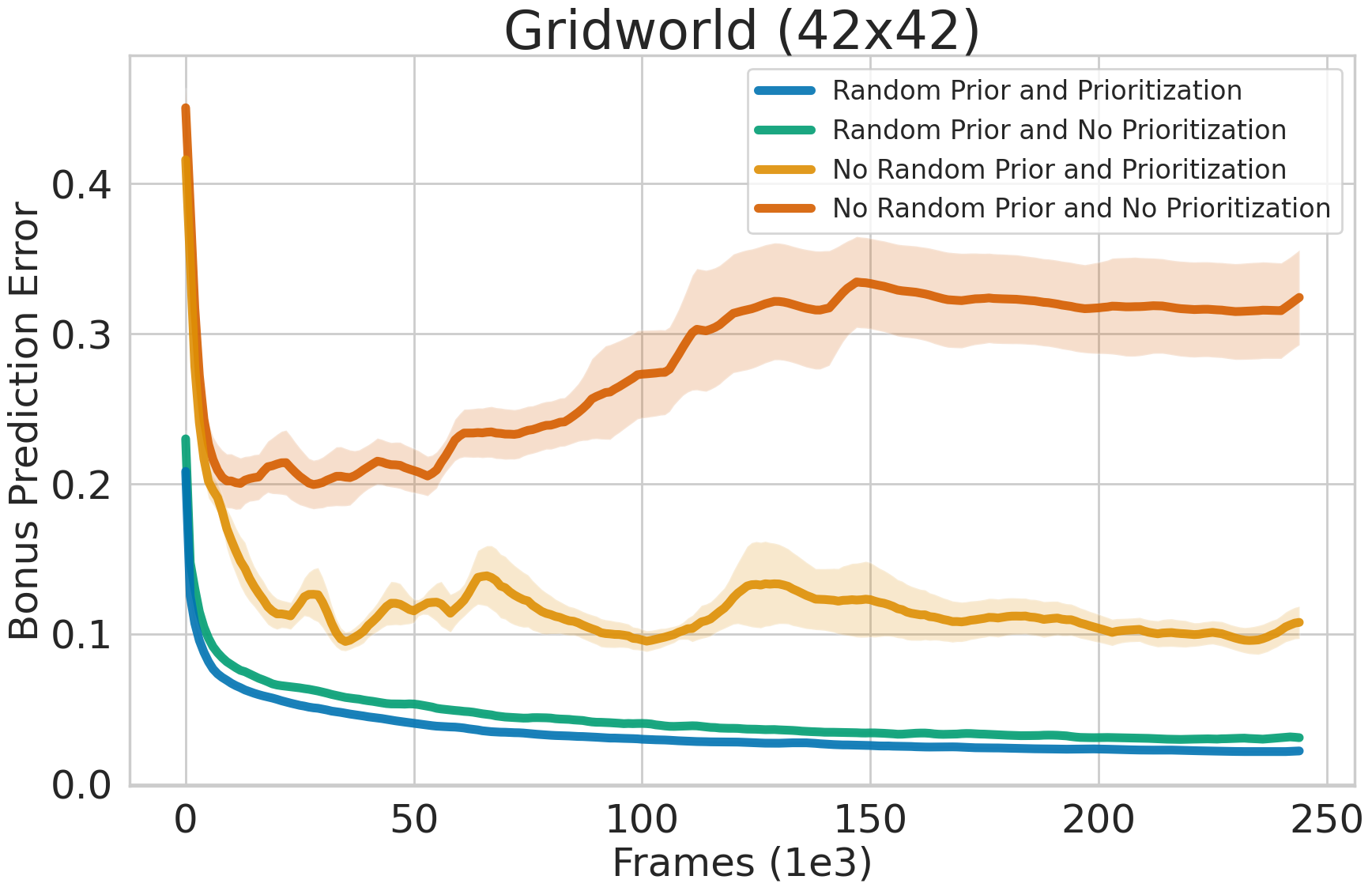}
    \vspace{-0.5cm}
    \caption{Ablating prioritized sampling and optimistic bonus initialization: vertical axis is the mean-squared error between the predicted and ground-truth count-based bonus (averaged over all visited states). Solid lines represent mean and bands represent standard error over $10$ random seeds; lower is better.}
    \vspace{-0.5cm}
    \label{fig:gridworld-ablation-mse}
\end{figure}

\begin{figure*}
    \centering
    \includegraphics[width=0.49\linewidth]{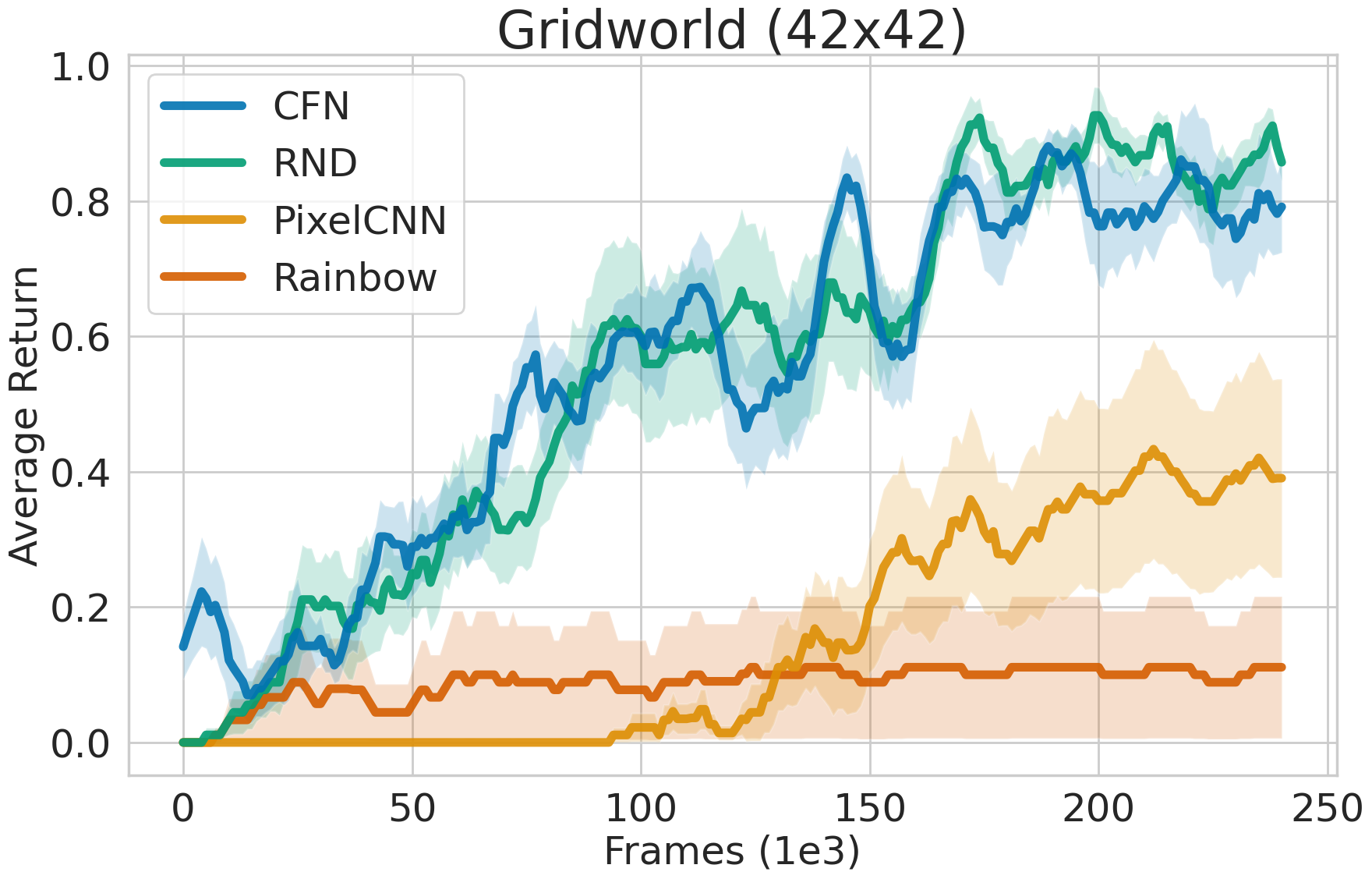}
    \hfill
    \includegraphics[width=0.49\linewidth]{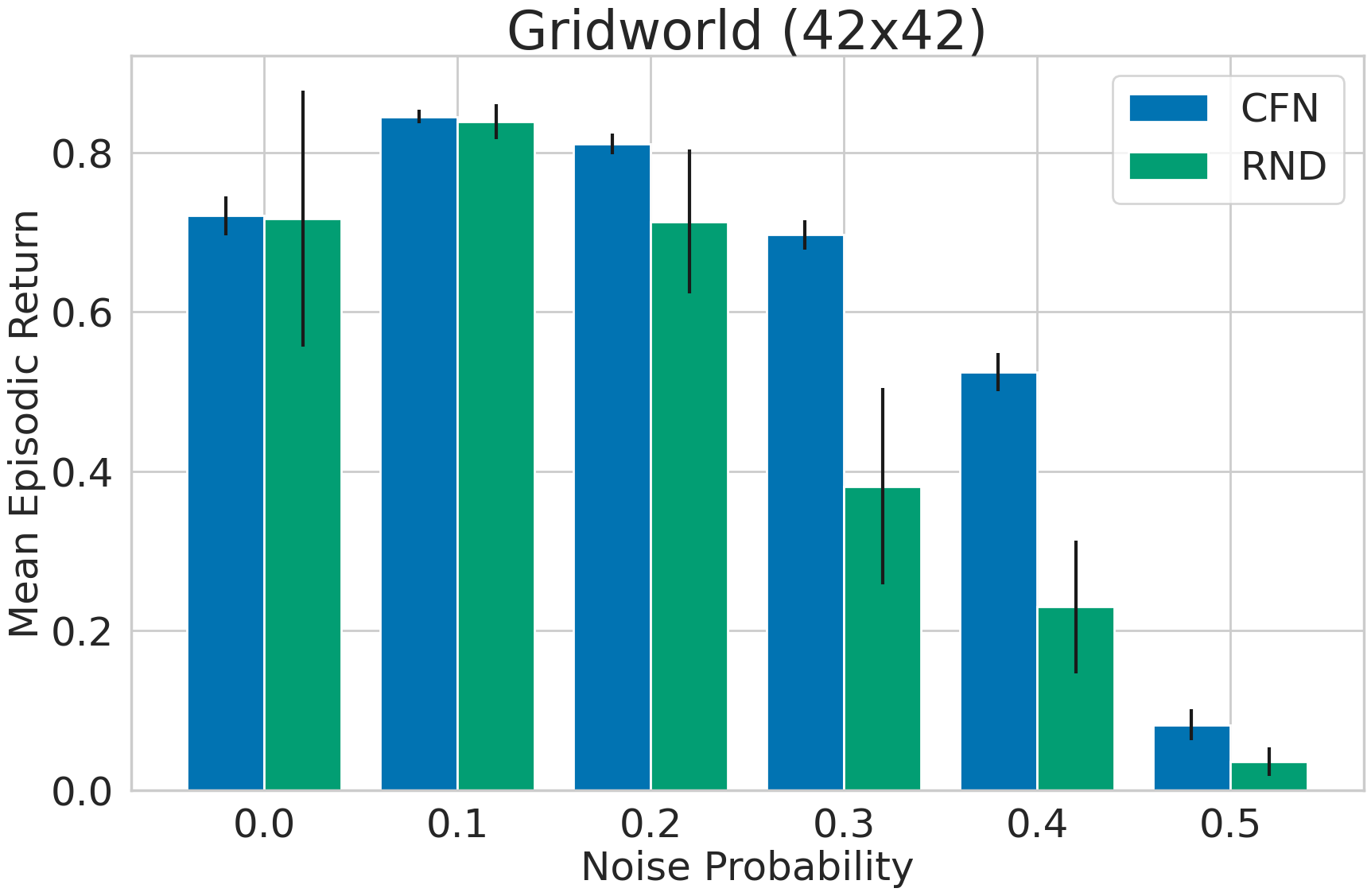}
    \caption{\textbf{Left}: Learning curves in deterministic Visual Gridworld comparing our method (CFN) with RND, PixelCNN and baseline Rainbow (with noisy networks). Solid lines denote mean episodic return, bands represent standard error.. \textbf{Right}: Comparison between CFN and RND on increasingly stochastic versions of Visual Gridworld. Bars represent mean episodic return averaged over training run, error bars denote standard error. All results are averaged over $10$ random seeds.}
    \label{fig:gridworld-learning-curves}
\end{figure*}

Our empirical results establish CFN as a competitive count-based exploration algorithm. First, we show that CFN can extract accurate counts in domains with visual observations, in contrast to other bonus methods. We then solve $8$ sparse-reward continuous control problems using CFN and show that we  significantly outperform RND and baseline SAC. Finally, we show that our method scales gracefully to the challenging Atari game \textsc{Montezuma's Revenge}.

\paragraph{Implementation details.} All exploration methods are built on top of Rainbow \citep{hessel2018rainbow} (when the action-space is discrete) or Soft Actor Critic (SAC) \citep{haarnoja2018soft} (when the action-space is continuous) using the Dopamine library \citep{castro18dopamine}. CFN has the same neural network architecture as RND's prediction network. We follow the experimental design of \citet{Taiga2020On}. We use a different set of hyperparameters for each \textit{suite} of tasks, i.e, one for Visual Gridworld, one for \textit{Fetch}, one for Ant, one for Adroit and one for Montezuma's Revenge. Details about CFN, including hyperparameters can be found in the Appendix~\ref{sec:appendix-hyperparameters}; details about environments can be found in Appendix~\ref{sec:appendix-environment-details}.
All code for reproducing results can be found at the linked repository.\footnote{https://github.com/samlobel/CFN}

\subsection{Bonus prediction accuracy}

We compare the exploration bonus from CFN to that of PixelCNN \citep{ostrovski2017count} and RND \citep{Burda2018} in Visual Gridworld \citep{allen2021learning}. Observations are $84$x$84$ images of a $42$x$42$ grid; these images serve as inputs during training. The agent is initialized in the bottom-left, and achieves a sparse terminal reward of $1$ for reaching the top-right within $150$ timesteps. For evaluation, we keep track of the tabular state and the ground-truth visitation counts.

Figure~\ref{fig:gridworld-counts} shows that while CFN is able to predict the count-based exploration bonus with high accuracy, PixelCNN and RND are not.  PixelCNN and CFN, being pseudocount methods, should ideally both output bonuses on the dashed line. Not only does PixelCNN mispredict the scale of the exploration bonus, it also assigns the same bonus to states visited once (${x=1)}$ versus those visited $25$ times (${x=0.2}$). RND's trendline is better than PixelCNN, although it has much higher variance than CFN. It is notable that for states with high count, its bonus falls off more sharply than $1/\sqrt{\mathcal{N}(s)}$. A similar experiment is repeated for the more challenging \textsc{Taxi} domain \citep{dietterich1998maxq} (with image observations); results are in Appendix~\ref{sec:appendix-taxi-counts}.

\subsection{Ablation: prioritization and random prior}\label{sec:experiments_ablation}
We now ablate the contribution of prioritized sampling (Section~\ref{sec:cfn_prioritization}) and random prior (Section~\ref{sec:cfn_randomprior}). In Figure~\ref{fig:gridworld-ablation-mse} we show how mean bonus prediction error evolves over time; the plot indicates that both additions lead to more accurate predictions. The prediction error is the mean-squared difference between the predicted and ground-truth exploration bonuses, averaged over unique states the agent has observed.

In Appendix~\ref{sec:appendix-per-rp-ablation} we provide more insight into each of these curves, and show how both these additions to CFN improve its bonus accuracy on states in the low-count regime. 

\subsection{RL performance on Visual Gridworld}\label{sec:experiements_gridworld_rl}
Figure~\ref{fig:gridworld-learning-curves} shows that CFN outperforms baseline Rainbow and PixelCNN, and performs similarly to RND on this task. An important feature to note about this environment is that it is deterministic. As such, the $1/\sqrt{\N(s)}$ bonus may not be appropriate because it is explicitly constructed to deal with stochastic environments \citep{auer2002using,jin2018q}.
So, we compare CFN to RND on a series of increasingly stochastic versions of Visual Gridworld.\footnote{Stochasticity is introduced by replacing the chosen action with a randomly selected action with some predefined probability.}
Our results show CFN's count-based bonus yields a more significant performance boost over RND in more stochastic versions of the problem. The slight performance bump at noise 0.1 can be attributed to the exploration benefit of random action-selection.

\subsection{Continuous Control Experiments}\label{sec:experiments_cc}

We now consider a series of challenging continuous control tasks. These are taken from two different suites: \textsc{Fetch} \citep{DBLP:journals/corr/abs-1802-09464} and \textsc{D4RL} \citep{fu2020d4rl}. For all tasks, we sparsify the reward function to make exploration challenging. We compare CFN to RND and baseline SAC; we do not compare against PixelCNN because the inputs are not images.

\begin{figure*}
    \centering
    \includegraphics[width=0.8\linewidth]{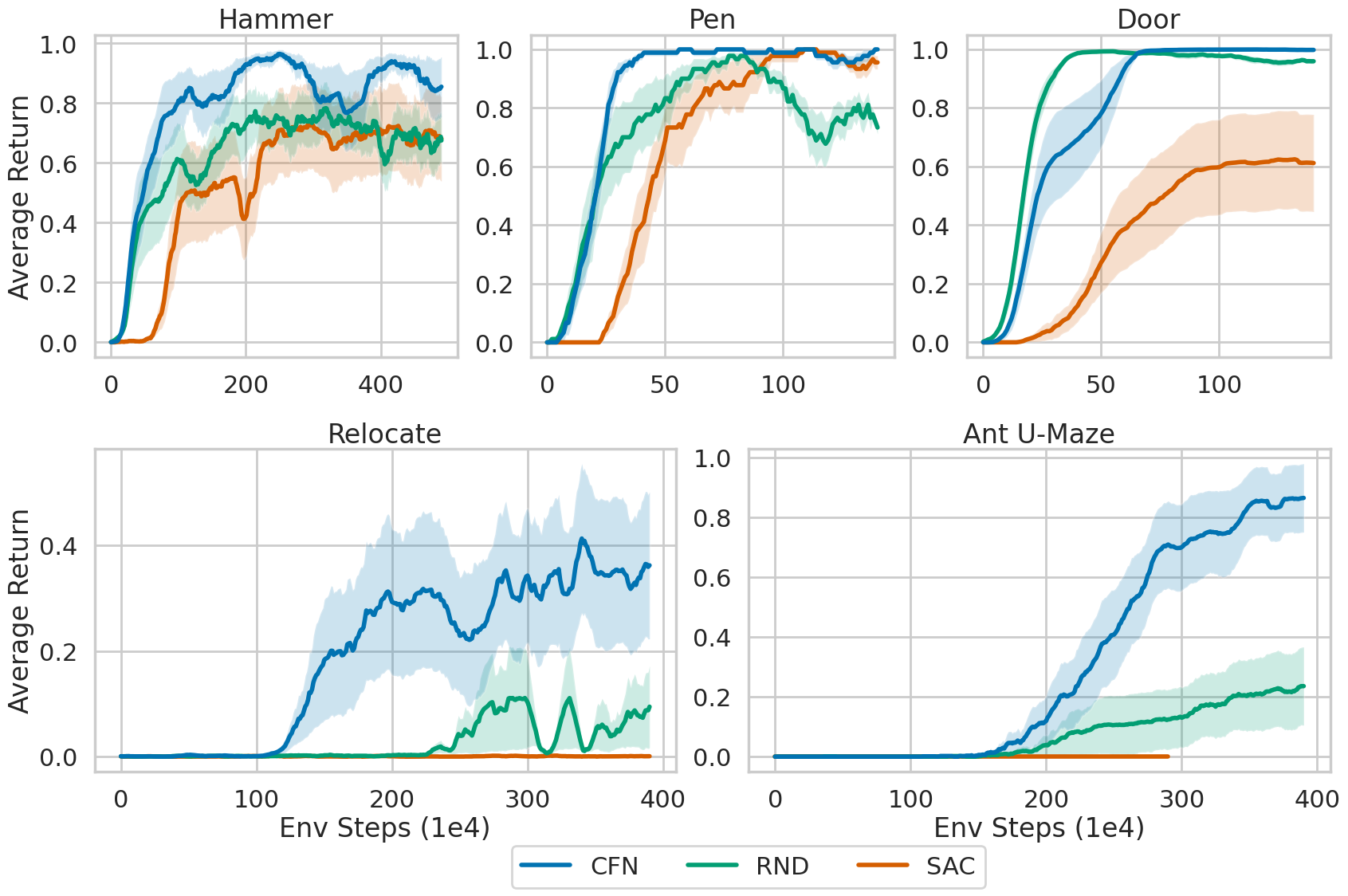}
    \caption{Learning curves in the D4RL tasks. Bottom row shows the $2$ most challenging tasks in this task suite. All curves are averaged over $9$ independent runs.}
    \label{fig:d4rl-all-learning-curves}
\end{figure*}

\begin{figure*}
    \centering
    \includegraphics[width=\linewidth]{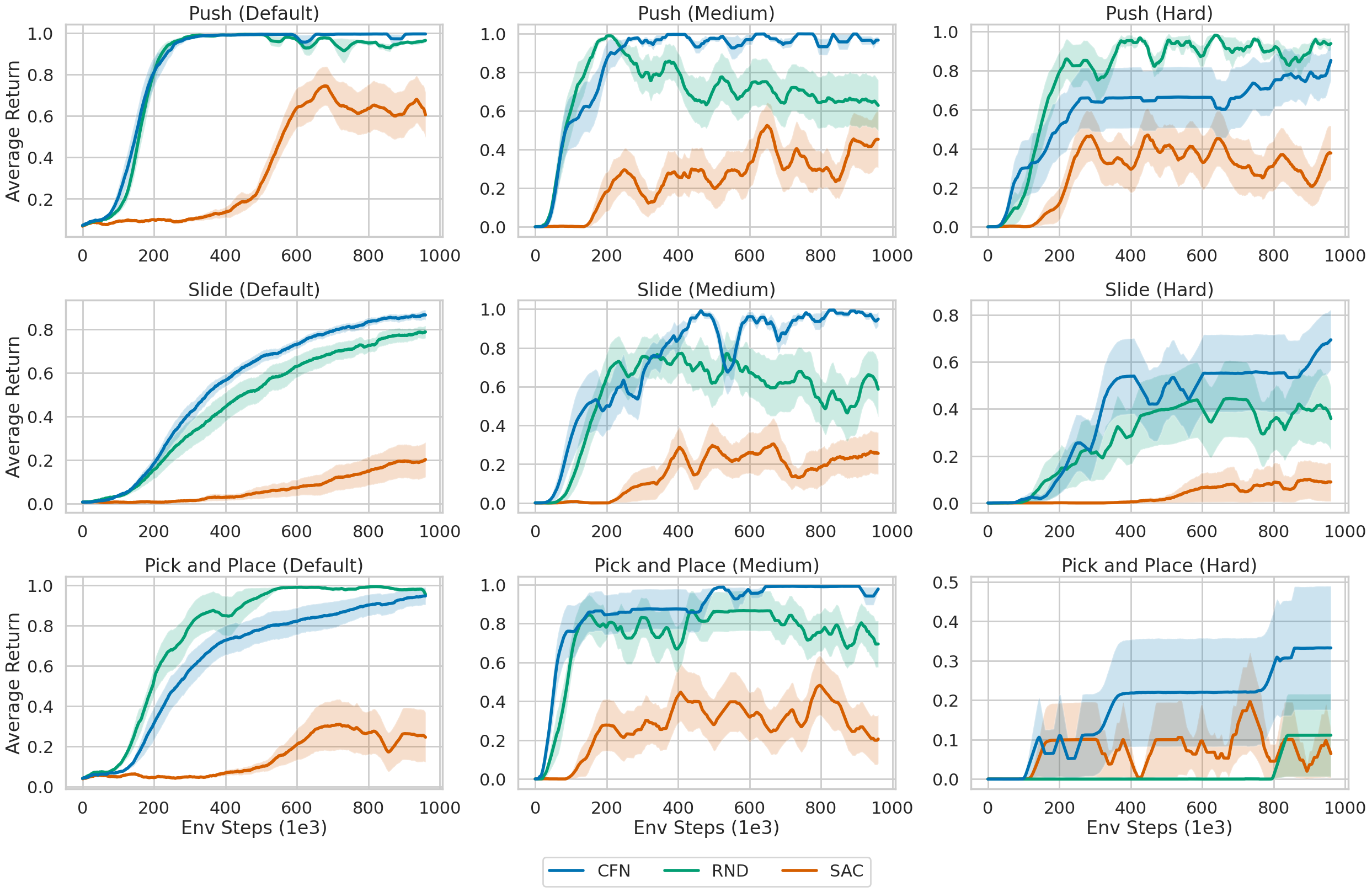}
    \vspace{-0.5cm}
    \caption{Results on the simulated \textsc{Fetch} manipulation tasks with $3$ levels of difficulty (default/easy, medium and hard). All curves are averaged over $9$ independent runs.}
    \label{fig:fetch-all-learning-curves}
\end{figure*}

\paragraph{Fetch manipulation tasks.} These tasks involve controlling a simulated Fetch robot to perform a series of manipulation tasks: pushing, sliding, or lifting an object to a goal location \citep{DBLP:journals/corr/abs-1802-09464}. We consider $3$ modes for each task: default, medium and hard; these modes differ in start-goal configurations. The default task randomizes the start-goal states (which occasionally exposes the agent to very simple episodes), medium and hard versions fix them to different levels of difficulty. Figure~\ref{fig:fetch-all-learning-curves} shows that both exploration methods outperform baseline SAC in all tasks; CFN outperforms RND on $6$ out of $9$ tasks and ties in $1$.

\paragraph{Ant-navigation and Adriot manipulation tasks.} Next, we consider tasks from D4RL.\citep{fu2020d4rl}\footnote{We use the domains from this suite, not their offline datasets.} The first involves controlling a quadrupedal ``ant'' robot in a U-shaped maze. The remaining $4$ tasks involve controlling a high-dimensional ``Adriot'' hand to perform various tasks: pick-and-place, reorienting a pen, opening a door and learning how to use a hammer \citep{Rajeswaran-RSS-18}. Similar to the Fetch tasks, we remove random restarts because they obviate the need for exploration \cite{lobel2022optimistic}. Figure~\ref{fig:d4rl-all-learning-curves} shows that CFN outperforms SAC on all tasks and RND on $4$ out of $5$ tasks.
More interestingly, the performance gains over RND are largest on the hardest exploration tasks (\textsc{Ant U-Maze} and \textsc{Relocate}; as evidenced by SAC's inability to experience any positive rewards). This supports the hypothesis that CFN provides a more thorough exploration bonus than RND.

\subsection{Performance in \textsc{Montezuma's Revenge}}\label{sec:monte_results}

\begin{figure*}
    \centering
    \includegraphics[width=0.49\linewidth]{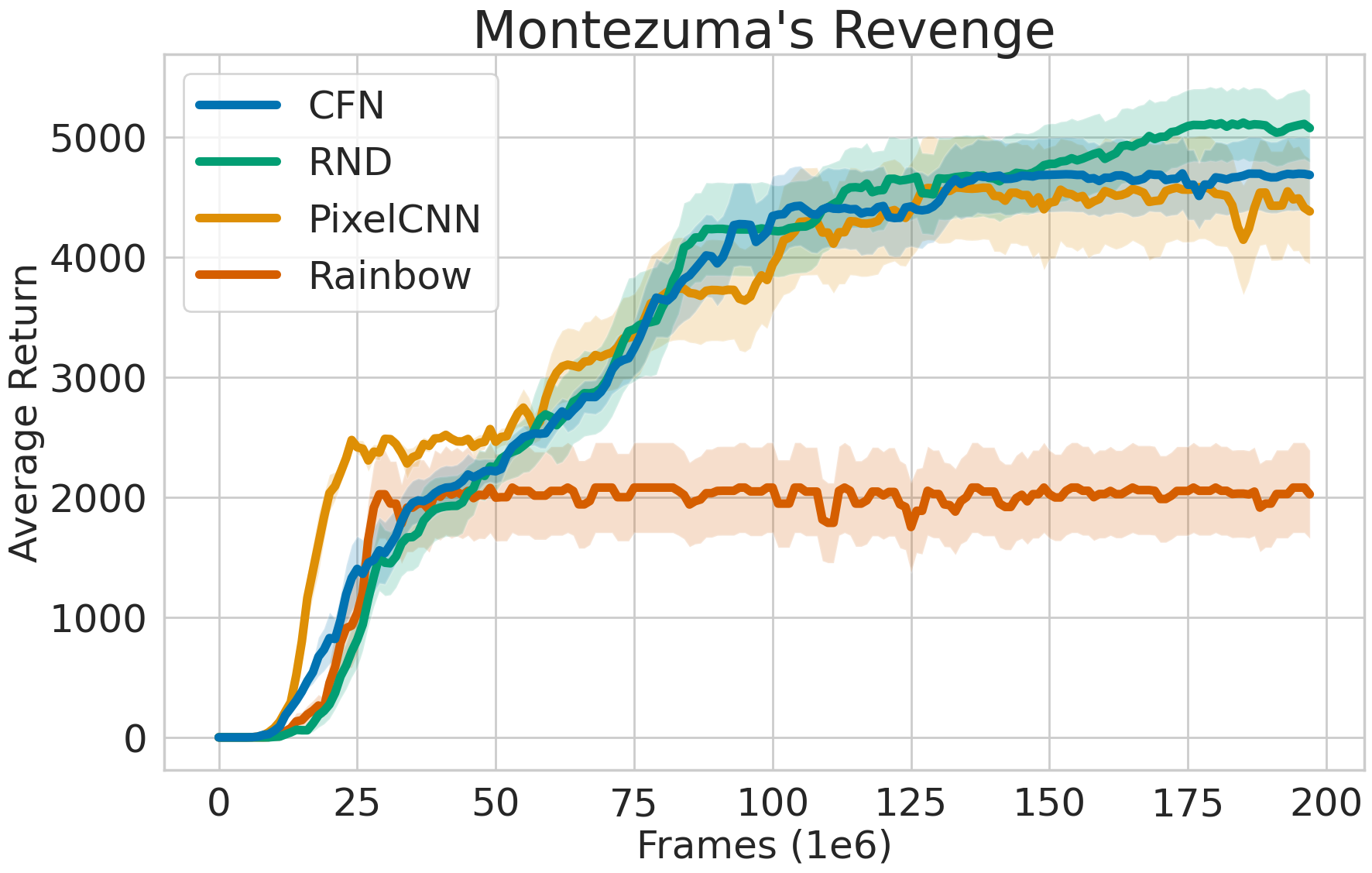}
    \hfill
    \includegraphics[width=0.49\linewidth]{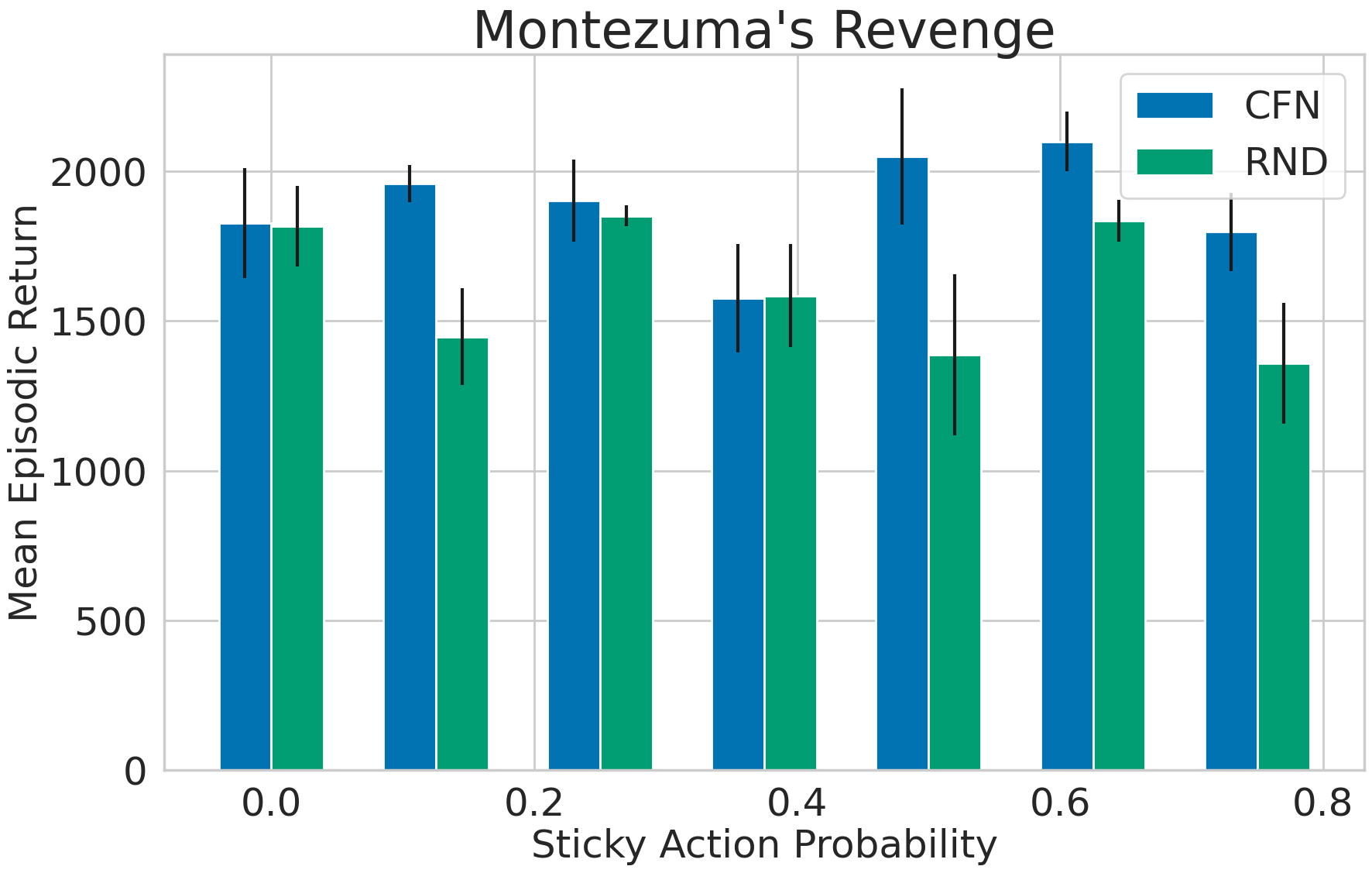}
    \caption{\textbf{Left}: Learning curve in Montezuma's Revenge comparing our method (CFN) with RND, PixelCNN and baseline Rainbow (with noisy networks). Solid lines denote mean episodic return, bands represent standard error averaged over $12$ random seeds. \textbf{Right}: Comparison between CFN and RND in terms of mean cumulative reward over $100$ million frames on versions of Montezuma's Revenge with varying ``sticky action'' probabilities (stochastic transitions; $0.25$ is the default sticky action probability \citep{machado2018revisiting}). Error bars represent standard error over 5 seeds.}
    \label{fig:monte-learning-curve}
\end{figure*}

Finally, we test our method on the challenging exploration benchmark: \textsc{Montezuma's Revenge}. We follow the experimental design suggested by \citet{machado2015domain} and compare CFN to baseline Rainbow, PixelCNN and RND.
Figure~\ref{fig:monte-learning-curve} shows that we comfortably outperform Rainbow in this task. All exploration algorithms perform similarly, a result also corroborated by \citet{Taiga2020On}.

Since all exploration methods perform similarly on the default task, we created a more challenging versions of \textsc{Montezuma's Revenge} by varying the amount of transition noise (via the ``sticky action'' probability \citep{machado2018revisiting}). Figure~\ref{fig:monte-learning-curve} \textit{(right)}
shows that CFN outperforms RND at higher levels of stochasticity;
this supports our hypothesis that count-based bonuses are better suited for stochastic environments than prediction-error based methods.

Notably, we find that having a large replay buffer for CFN slightly improves performance, which increases memory requirements for this experiment. More discussion about the impact of buffer size can be found in Appendix~\ref{sec:appendix-cfn-replay-size}; details about hyperparameters can be found in Appendix~\ref{sec:appendix-hyperparameters}.

\section{Conclusion and Future Work}
Though count-based exploration is a principled way to do exploration, it is not the dominant approach used in practice; we aim to remedy that. As a step in that direction, we presented a new method for count-based exploration which extracts pseudocounts by learning to average samples from the Rademacher distribution. In contrast to prior pseudocount methods, ours produces accurate counts in simple problems and can be flexibly applied to a variety of observation spaces. We demonstrate strong results on \textsc{Montezuma's Revenge} and $8$ challenging continuous control problems. Directions for future work include the use of representation learning techniques that capture MDP-specific structure \citep{allen2021learning}, incorporating actions into the exploration bonus, as well as a mechanism for ``forgetting'' high-count states from the replay buffer.

\section*{Acknowledgements}
We would like to thank Georg Ostrovski for his guidance, Adrien Ali Taiga for pointing us to the bonus-based exploration code, and Cameron Allen for help with the Visual Gridworld and Taxi domain implementations.
Part of this research was conducted using computational resources and services at the Center for Computation and Visualization, Brown University.
This research was supported in part by NSF grant \#1955361,
NSF CAREER award \#1844960, NSF GRFP award \#2040433, and an Amazon Research Award.

\bibliography{example_paper}

\begin{thebibliography}{55}
\providecommand{\natexlab}[1]{#1}
\providecommand{\url}[1]{\texttt{#1}}
\expandafter\ifx\csname urlstyle\endcsname\relax
  \providecommand{\doi}[1]{doi: #1}\else
  \providecommand{\doi}{doi: \begingroup \urlstyle{rm}\Url}\fi

\bibitem[Allen et~al.(2021)Allen, Parikh, Gottesman, and
  Konidaris]{allen2021learning}
Allen, C., Parikh, N., Gottesman, O., and Konidaris, G.
\newblock Learning markov state abstractions for deep reinforcement learning.
\newblock \emph{Advances in Neural Information Processing Systems}, 34, 2021.

\bibitem[Auer(2002)]{auer2002using}
Auer, P.
\newblock Using confidence bounds for exploitation-exploration trade-offs.
\newblock \emph{Journal of Machine Learning Research}, 3\penalty0
  (Nov):\penalty0 397--422, 2002.

\bibitem[Azar et~al.(2017)Azar, Osband, and Munos]{mohammad2017minimax}
Azar, M.~G., Osband, I., and Munos, R.
\newblock Minimax regret bounds for reinforcement learning.
\newblock In \emph{Proceedings of the 34th International Conference on Machine
  Learning}, volume~70 of \emph{Proceedings of Machine Learning Research}, pp.\
   263--272. PMLR, 06--11 Aug 2017.
\newblock URL \url{https://proceedings.mlr.press/v70/azar17a.html}.

\bibitem[Badia et~al.(2020{\natexlab{a}})Badia, Piot, Kapturowski, Sprechmann,
  Vitvitskyi, Guo, and Blundell]{badia2020agent57}
Badia, A.~P., Piot, B., Kapturowski, S., Sprechmann, P., Vitvitskyi, A., Guo,
  Z.~D., and Blundell, C.
\newblock Agent57: Outperforming the atari human benchmark.
\newblock In \emph{International Conference on Machine Learning}, pp.\
  507--517. PMLR, 2020{\natexlab{a}}.

\bibitem[Badia et~al.(2020{\natexlab{b}})Badia, Sprechmann, Vitvitskyi, Guo,
  Piot, Kapturowski, Tieleman, Arjovsky, Pritzel, Bolt, et~al.]{badia2020never}
Badia, A.~P., Sprechmann, P., Vitvitskyi, A., Guo, D., Piot, B., Kapturowski,
  S., Tieleman, O., Arjovsky, M., Pritzel, A., Bolt, A., et~al.
\newblock Never give up: Learning directed exploration strategies.
\newblock \emph{arXiv preprint arXiv:2002.06038}, 2020{\natexlab{b}}.

\bibitem[Bagaria et~al.(2021)Bagaria, Senthil, and Konidaris]{bagaria2021skill}
Bagaria, A., Senthil, J.~K., and Konidaris, G.
\newblock Skill discovery for exploration and planning using deep skill graphs.
\newblock In \emph{International Conference on Machine Learning}, pp.\
  521--531. PMLR, 2021.

\bibitem[Bellemare et~al.(2016)Bellemare, Srinivasan, Ostrovski, Schaul,
  Saxton, and Munos]{bellemare2016unifying}
Bellemare, M., Srinivasan, S., Ostrovski, G., Schaul, T., Saxton, D., and
  Munos, R.
\newblock Unifying count-based exploration and intrinsic motivation.
\newblock In \emph{Advances in Neural Information Processing Systems}, pp.\
  1471--1479, 2016.

\bibitem[Bellemare et~al.(2013)Bellemare, Naddaf, Veness, and
  Bowling]{bellemare2013arcade}
Bellemare, M.~G., Naddaf, Y., Veness, J., and Bowling, M.
\newblock The arcade learning environment: An evaluation platform for general
  agents.
\newblock \emph{Journal of Artificial Intelligence Research}, 47:\penalty0
  253--279, 2013.

\bibitem[Brafman \& Tennenholtz(2002)Brafman and Tennenholtz]{brafman2002r}
Brafman, R.~I. and Tennenholtz, M.
\newblock R-max-a general polynomial time algorithm for near-optimal
  reinforcement learning.
\newblock \emph{Journal of Machine Learning Research}, 3\penalty0
  (Oct):\penalty0 213--231, 2002.

\bibitem[Burda et~al.(2018)Burda, Edwards, Pathak, Storkey, Darrell, and
  Efros]{burda2018large}
Burda, Y., Edwards, H., Pathak, D., Storkey, A., Darrell, T., and Efros, A.~A.
\newblock Large-scale study of curiosity-driven learning.
\newblock In \emph{International Conference on Learning Representations}, 2018.

\bibitem[Burda et~al.(2019)Burda, Edwards, Storkey, and Klimov]{Burda2018}
Burda, Y., Edwards, H., Storkey, A., and Klimov, O.
\newblock Exploration by random network distillation.
\newblock In \emph{International Conference on Learning Representations}, 2019.
\newblock URL \url{https://openreview.net/forum?id=H1lJJnR5Ym}.

\bibitem[Castro et~al.(2018)Castro, Moitra, Gelada, Kumar, and
  Bellemare]{castro18dopamine}
Castro, P.~S., Moitra, S., Gelada, C., Kumar, S., and Bellemare, M.~G.
\newblock Dopamine: {A} {R}esearch {F}ramework for {D}eep {R}einforcement
  {L}earning.
\newblock 2018.
\newblock URL \url{http://arxiv.org/abs/1812.06110}.

\bibitem[Dayan(1993)]{dayan1993improving}
Dayan, P.
\newblock Improving generalization for temporal difference learning: The
  successor representation.
\newblock \emph{Neural computation}, 5\penalty0 (4):\penalty0 613--624, 1993.

\bibitem[Dietterich(1998)]{dietterich1998maxq}
Dietterich, T.
\newblock The {MAXQ} method for hierarchical reinforcement learning.
\newblock In \emph{ICML}, volume~98, pp.\  118--126, 1998.

\bibitem[Diuk et~al.(2008)Diuk, Cohen, and Littman]{diuk2008object}
Diuk, C., Cohen, A., and Littman, M.~L.
\newblock An object-oriented representation for efficient reinforcement
  learning.
\newblock In \emph{Proceedings of the 25th international conference on Machine
  learning}, pp.\  240--247, 2008.

\bibitem[Ecoffet et~al.(2021)Ecoffet, Huizinga, Lehman, Stanley, and
  Clune]{ecoffet2021first}
Ecoffet, A., Huizinga, J., Lehman, J., Stanley, K.~O., and Clune, J.
\newblock First return, then explore.
\newblock \emph{Nature}, 590\penalty0 (7847):\penalty0 580--586, 2021.

\bibitem[Ermolov \& Sebe(2020)Ermolov and Sebe]{ermolov2020latent}
Ermolov, A. and Sebe, N.
\newblock Latent world models for intrinsically motivated exploration.
\newblock \emph{Advances in Neural Information Processing Systems},
  33:\penalty0 5565--5575, 2020.

\bibitem[Fu et~al.(2020)Fu, Kumar, Nachum, Tucker, and Levine]{fu2020d4rl}
Fu, J., Kumar, A., Nachum, O., Tucker, G., and Levine, S.
\newblock D4rl: Datasets for deep data-driven reinforcement learning.
\newblock \emph{arXiv preprint arXiv:2004.07219}, 2020.

\bibitem[Golub \& Reinsch(1971)Golub and Reinsch]{golub1971singular}
Golub, G.~H. and Reinsch, C.
\newblock Singular value decomposition and least squares solutions.
\newblock \emph{Linear algebra}, 2:\penalty0 134--151, 1971.

\bibitem[Guo et~al.(2022)Guo, Thakoor, P{\^\i}slar, Pires, Altch{\'e}, Tallec,
  Saade, Calandriello, Grill, Tang, et~al.]{guo2022byol}
Guo, Z.~D., Thakoor, S., P{\^\i}slar, M., Pires, B.~A., Altch{\'e}, F., Tallec,
  C., Saade, A., Calandriello, D., Grill, J.-B., Tang, Y., et~al.
\newblock Byol-explore: Exploration by bootstrapped prediction.
\newblock \emph{arXiv preprint arXiv:2206.08332}, 2022.

\bibitem[Haarnoja et~al.(2018)Haarnoja, Zhou, Abbeel, and
  Levine]{haarnoja2018soft}
Haarnoja, T., Zhou, A., Abbeel, P., and Levine, S.
\newblock Soft actor-critic: Off-policy maximum entropy deep reinforcement
  learning with a stochastic actor.
\newblock In \emph{International conference on machine learning}, pp.\
  1861--1870. PMLR, 2018.

\bibitem[Hessel et~al.(2018)Hessel, Modayil, Van~Hasselt, Schaul, Ostrovski,
  Dabney, Horgan, Piot, Azar, and Silver]{hessel2018rainbow}
Hessel, M., Modayil, J., Van~Hasselt, H., Schaul, T., Ostrovski, G., Dabney,
  W., Horgan, D., Piot, B., Azar, M., and Silver, D.
\newblock Rainbow: Combining improvements in deep reinforcement learning.
\newblock In \emph{Thirty-second AAAI conference on artificial intelligence},
  2018.

\bibitem[Houthooft et~al.(2016)Houthooft, Chen, Duan, Schulman, De~Turck, and
  Abbeel]{houthooft2016vime}
Houthooft, R., Chen, X., Duan, Y., Schulman, J., De~Turck, F., and Abbeel, P.
\newblock Vime: Variational information maximizing exploration.
\newblock \emph{Advances in neural information processing systems}, 29, 2016.

\bibitem[Jin et~al.(2018)Jin, Allen-Zhu, Bubeck, and Jordan]{jin2018q}
Jin, C., Allen-Zhu, Z., Bubeck, S., and Jordan, M.~I.
\newblock Is q-learning provably efficient?
\newblock \emph{Advances in neural information processing systems}, 31, 2018.

\bibitem[Kakade et~al.(2003)Kakade, Kearns, and
  Langford]{kakade2003exploration}
Kakade, S., Kearns, M.~J., and Langford, J.
\newblock Exploration in metric state spaces.
\newblock In \emph{Proceedings of the 20th International Conference on Machine
  Learning (ICML-03)}, pp.\  306--312, 2003.

\bibitem[Kapturowski et~al.(2022)Kapturowski, Campos, Jiang, Raki{\'c}evi{\'c},
  van Hasselt, Blundell, and Badia]{kapturowski2022human}
Kapturowski, S., Campos, V., Jiang, R., Raki{\'c}evi{\'c}, N., van Hasselt, H.,
  Blundell, C., and Badia, A.~P.
\newblock Human-level atari 200x faster.
\newblock \emph{arXiv preprint arXiv:2209.07550}, 2022.

\bibitem[Kearns \& Singh(2002)Kearns and Singh]{kearns2002near}
Kearns, M. and Singh, S.
\newblock Near-optimal reinforcement learning in polynomial time.
\newblock \emph{Machine learning}, 49\penalty0 (2):\penalty0 209--232, 2002.

\bibitem[Lillicrap et~al.(2015)Lillicrap, Hunt, Pritzel, Heess, Erez, Tassa,
  Silver, and Wierstra]{lillicrap2015continuous}
Lillicrap, T.~P., Hunt, J.~J., Pritzel, A., Heess, N., Erez, T., Tassa, Y.,
  Silver, D., and Wierstra, D.
\newblock Continuous control with deep reinforcement learning.
\newblock \emph{arXiv preprint arXiv:1509.02971}, 2015.

\bibitem[Lobel et~al.(2022)Lobel, Gottesman, Allen, Bagaria, and
  Konidaris]{lobel2022optimistic}
Lobel, S., Gottesman, O., Allen, C., Bagaria, A., and Konidaris, G.
\newblock Optimistic initialization for exploration in continuous control.
\newblock In \emph{Proceedings of the AAAI Conference on Artificial
  Intelligence}, volume~36, pp.\  7612--7619, 2022.

\bibitem[Machado et~al.(2015)Machado, Srinivasan, and
  Bowling]{machado2015domain}
Machado, M.~C., Srinivasan, S., and Bowling, M.
\newblock Domain-independent optimistic initialization for reinforcement
  learning.
\newblock In \emph{Workshops at the Twenty-Ninth AAAI Conference on Artificial
  Intelligence}, 2015.

\bibitem[Machado et~al.(2018)Machado, Bellemare, Talvitie, Veness, Hausknecht,
  and Bowling]{machado2018revisiting}
Machado, M.~C., Bellemare, M.~G., Talvitie, E., Veness, J., Hausknecht, M., and
  Bowling, M.
\newblock Revisiting the arcade learning environment: Evaluation protocols and
  open problems for general agents.
\newblock \emph{Journal of Artificial Intelligence Research}, 61:\penalty0
  523--562, 2018.

\bibitem[Machado et~al.(2020)Machado, Bellemare, and Bowling]{machado2020count}
Machado, M.~C., Bellemare, M.~G., and Bowling, M.
\newblock Count-based exploration with the successor representation.
\newblock In \emph{Proceedings of the AAAI Conference on Artificial
  Intelligence}, volume~34, pp.\  5125--5133, 2020.

\bibitem[Mnih et~al.(2015)Mnih, Kavukcuoglu, Silver, Rusu, Veness, Bellemare,
  Graves, Riedmiller, Fidjeland, Ostrovski, et~al.]{mnih2015human}
Mnih, V., Kavukcuoglu, K., Silver, D., Rusu, A.~A., Veness, J., Bellemare,
  M.~G., Graves, A., Riedmiller, M., Fidjeland, A.~K., Ostrovski, G., et~al.
\newblock Human-level control through deep reinforcement learning.
\newblock \emph{Nature}, 518\penalty0 (7540):\penalty0 529--533, 2015.

\bibitem[Osband et~al.(2016)Osband, Blundell, Pritzel, and
  Van~Roy]{osband2016bootstrap}
Osband, I., Blundell, C., Pritzel, A., and Van~Roy, B.
\newblock Deep exploration via bootstrapped dqn.
\newblock In Lee, D., Sugiyama, M., Luxburg, U., Guyon, I., and Garnett, R.
  (eds.), \emph{Advances in Neural Information Processing Systems}, volume~29.
  Curran Associates, Inc., 2016.
\newblock URL
  \url{https://proceedings.neurips.cc/paper/2016/file/8d8818c8e140c64c743113f563cf750f-Paper.pdf}.

\bibitem[Osband et~al.(2018)Osband, Aslanides, and
  Cassirer]{osband2018randomized}
Osband, I., Aslanides, J., and Cassirer, A.
\newblock Randomized prior functions for deep reinforcement learning.
\newblock \emph{Advances in Neural Information Processing Systems}, 31, 2018.

\bibitem[Ostrovski et~al.(2017)Ostrovski, Bellemare, Oord, and
  Munos]{ostrovski2017count}
Ostrovski, G., Bellemare, M.~G., Oord, A., and Munos, R.
\newblock Count-based exploration with neural density models.
\newblock In \emph{International conference on machine learning}, pp.\
  2721--2730. PMLR, 2017.

\bibitem[O’Donoghue et~al.(2018)O’Donoghue, Osband, Munos, and
  Mnih]{o2018uncertainty}
O’Donoghue, B., Osband, I., Munos, R., and Mnih, V.
\newblock The uncertainty bellman equation and exploration.
\newblock In \emph{International Conference on Machine Learning}, pp.\
  3836--3845, 2018.

\bibitem[Pathak et~al.(2017)Pathak, Agrawal, Efros, and
  Darrell]{pathak2017curiosity}
Pathak, D., Agrawal, P., Efros, A.~A., and Darrell, T.
\newblock Curiosity-driven exploration by self-supervised prediction.
\newblock In \emph{International Conference on Machine Learning}, pp.\
  2778--2787. PMLR, 2017.

\bibitem[Pitis et~al.(2020)Pitis, Chan, Zhao, Stadie, and Ba]{pitis2020maximum}
Pitis, S., Chan, H., Zhao, S., Stadie, B., and Ba, J.
\newblock Maximum entropy gain exploration for long horizon multi-goal
  reinforcement learning.
\newblock \emph{arXiv preprint arXiv:2007.02832}, 2020.

\bibitem[Plappert et~al.(2018)Plappert, Andrychowicz, Ray, McGrew, Baker,
  Powell, Schneider, Tobin, Chociej, Welinder, Kumar, and
  Zaremba]{DBLP:journals/corr/abs-1802-09464}
Plappert, M., Andrychowicz, M., Ray, A., McGrew, B., Baker, B., Powell, G.,
  Schneider, J., Tobin, J., Chociej, M., Welinder, P., Kumar, V., and Zaremba,
  W.
\newblock Multi-goal reinforcement learning: Challenging robotics environments
  and request for research.
\newblock \emph{CoRR}, abs/1802.09464, 2018.
\newblock URL \url{http://arxiv.org/abs/1802.09464}.

\bibitem[Pong et~al.(2019)Pong, Dalal, Lin, Nair, Bahl, and
  Levine]{pong2019skew}
Pong, V.~H., Dalal, M., Lin, S., Nair, A., Bahl, S., and Levine, S.
\newblock Skew-fit: State-covering self-supervised reinforcement learning.
\newblock \emph{Proceedings of the 37th International Conference on Machine
  Learning, {ICML}}, 2019.

\bibitem[Raileanu \& Rockt{\"a}schel(2020)Raileanu and
  Rockt{\"a}schel]{raileanu2020ride}
Raileanu, R. and Rockt{\"a}schel, T.
\newblock Ride: Rewarding impact-driven exploration for procedurally-generated
  environments.
\newblock \emph{arXiv preprint arXiv:2002.12292}, 2020.

\bibitem[Rajeswaran et~al.(2018)Rajeswaran, Kumar, Gupta, Vezzani, Schulman,
  Todorov, and Levine]{Rajeswaran-RSS-18}
Rajeswaran, A., Kumar, V., Gupta, A., Vezzani, G., Schulman, J., Todorov, E.,
  and Levine, S.
\newblock {Learning Complex Dexterous Manipulation with Deep Reinforcement
  Learning and Demonstrations}.
\newblock In \emph{Proceedings of Robotics: Science and Systems (RSS)}, 2018.

\bibitem[Rashid et~al.(2020)Rashid, Peng, Boehmer, and
  Whiteson]{Rashid2020Optimistic}
Rashid, T., Peng, B., Boehmer, W., and Whiteson, S.
\newblock Optimistic exploration even with a pessimistic initialisation.
\newblock In \emph{International Conference on Learning Representations}, 2020.
\newblock URL \url{https://openreview.net/forum?id=r1xGP6VYwH}.

\bibitem[Schaul et~al.(2015)Schaul, Quan, Antonoglou, and
  Silver]{schaul2015prioritized}
Schaul, T., Quan, J., Antonoglou, I., and Silver, D.
\newblock Prioritized experience replay.
\newblock \emph{arXiv preprint arXiv:1511.05952}, 2015.

\bibitem[Stadie et~al.(2015)Stadie, Levine, and
  Abbeel]{stadie2015incentivizing}
Stadie, B.~C., Levine, S., and Abbeel, P.
\newblock Incentivizing exploration in reinforcement learning with deep
  predictive models.
\newblock \emph{arXiv preprint arXiv:1507.00814}, 2015.

\bibitem[Strehl \& Littman(2008)Strehl and Littman]{strehl2008analysis}
Strehl, A.~L. and Littman, M.~L.
\newblock An analysis of model-based interval estimation for markov decision
  processes.
\newblock \emph{Journal of Computer and System Sciences}, 74\penalty0
  (8):\penalty0 1309--1331, 2008.

\bibitem[Sutton \& Barto(2018)Sutton and Barto]{sutton2018reinforcement}
Sutton, R.~S. and Barto, A.~G.
\newblock \emph{Reinforcement learning: An introduction}.
\newblock MIT press, 2018.

\bibitem[Sutton et~al.(2022)Sutton, Bowling, and Pilarski]{sutton2022alberta}
Sutton, R.~S., Bowling, M.~H., and Pilarski, P.~M.
\newblock The {A}lberta plan for {AI} research.
\newblock \emph{arXiv preprint arXiv:2208.11173}, 2022.

\bibitem[Taiga et~al.(2020)Taiga, Fedus, Machado, Courville, and
  Bellemare]{Taiga2020On}
Taiga, A.~A., Fedus, W., Machado, M.~C., Courville, A., and Bellemare, M.~G.
\newblock On bonus based exploration methods in the arcade learning
  environment.
\newblock In \emph{International Conference on Learning Representations}, 2020.
\newblock URL \url{https://openreview.net/forum?id=BJewlyStDr}.

\bibitem[Tang et~al.(2017)Tang, Houthooft, Foote, Stooke, Xi~Chen, Duan,
  Schulman, DeTurck, and Abbeel]{tang2017exploration}
Tang, H., Houthooft, R., Foote, D., Stooke, A., Xi~Chen, O., Duan, Y.,
  Schulman, J., DeTurck, F., and Abbeel, P.
\newblock \# exploration: A study of count-based exploration for deep
  reinforcement learning.
\newblock \emph{Advances in neural information processing systems}, 30, 2017.

\bibitem[Thompson(1933)]{thompson1933likelihood}
Thompson, W.~R.
\newblock On the likelihood that one unknown probability exceeds another in
  view of the evidence of two samples.
\newblock \emph{Biometrika}, 25\penalty0 (3-4):\penalty0 285--294, 1933.

\bibitem[Watkins \& Dayan(1992)Watkins and Dayan]{watkins1992q}
Watkins, C.~J. and Dayan, P.
\newblock Q-learning.
\newblock \emph{Machine learning}, 8\penalty0 (3-4):\penalty0 279--292, 1992.

\bibitem[Zhang \& Sutton(2017)Zhang and Sutton]{zhang2017deeper}
Zhang, S. and Sutton, R.~S.
\newblock A deeper look at experience replay.
\newblock \emph{arXiv preprint arXiv:1712.01275}, 2017.

\bibitem[Zhang et~al.(2021)Zhang, Xu, Wang, Wu, Keutzer, Gonzalez, and
  Tian]{zhang2021noveld}
Zhang, T., Xu, H., Wang, X., Wu, Y., Keutzer, K., Gonzalez, J.~E., and Tian, Y.
\newblock Noveld: A simple yet effective exploration criterion.
\newblock \emph{Advances in Neural Information Processing Systems},
  34:\penalty0 25217--25230, 2021.

\end{thebibliography}
\bibliographystyle{icml2023}

\newpage
\appendix
\onecolumn


\section{Proofs}\label{sec:appendix-proofs}
\subsection{Any zero-mean unit-variance distribution can be used for counting} \label{sec:appendix-any-whitened-distribution-counts}
Let $z_n = \frac{1}{n} \sum_{i=1}^n x_i$ where $x_i \sim \mathcal{X}$. Assume that the distribution $\mathcal{X}$ is such that $\mathbb{E}[\mathcal{X}] = 0$ and $\text{Var}[\mathcal{X}] = 1$. In many cases we make use of the fact that $\mathbb{E}[x_ix_j] = \delta_{ij}$ (Kronecker delta function), because if $i\neq j$, then $\mathbb{E}[x_ix_j] = \mathbb{E}[x_i]\mathbb{E}[x_j] = 0$

\begin{align*}
\mathbb{E}\big[z^2_n\big]
&= \mathbb{E}\Big[\Big(\frac{1}{n}\sum_{i=1}^n x_i\Big)^2\Big]\\
&=\frac{1}{n^2}\mathbb{E}\Big[\sum_{i=1}^n\sum_{j=1}^n x_i x_j\Big]\\
&=\frac{1}{n^2}\mathbb{E}\Big[\sum_{i=1}^n x_i x_i + \sum_{i=1}^n \sum_{j\neq i}^n x_i x_j\Big]\\
&=\frac{1}{n^2}\mathbb{E}\Big[\sum_{i=1}^n x_i^2\Big] + \frac{1}{n^2}\mathbb{E}\Big[\sum_{i=1}^n \sum_{j\neq i}^n x_i x_j\Big]\\
&=\frac{1}{n^2}(n + 0)\\
&=\frac{1}{n}\\
\end{align*}

This proves the well-known fact that the variance of the sample mean scales inversely with the number of samples.

\subsection{Functional form of the variance of $z_n^2$}\label{sec:appendix-variance-of-inverse-count-estimator}
We now know that $z^2_n$ is an unbiased estimator of $\frac{1}{n}$, the inverse count. What is the variance of this estimate?
\begin{align*}
    \text{Var}\big[z^2_n\big]
    &= \mathbb{E}\big[z_n^4\big] - \mathbb{E}\big[z_n^2\big]^2\\
    &= \mathbb{E}\Big[(\frac{1}{n}\sum_{i=1}^n x_i)^4\Big] - \frac{1}{n^2}\\
    &= \frac{1}{n^4}\mathbb{E}\Big[(\sum_{i=1}^n x_i)^4\Big] - \frac{1}{n^2}\\
    \mathbb{E}\Big[(\sum_{i=1}^n x_i)^4\Big] &= \mathbb{E}\Big[\sum_i\sum_j\sum_k\sum_l x_ix_jx_kx_l\Big]\\
    &= \mathbb{E}\Big[\sum_{i=j=k=l} x_i^4 + \sum_{i=j,k=l\neq i}x_i^2 x_k^2 + \sum_{i=k,l=j\neq i}x_i^2 x_l^2 + \sum_{i=l,j=k\neq i}x_i^2 x_k^2 + \sum_{\text{remaining}}x_ix_jx_kx_l\Big]
\end{align*}
where here we have broken the summation across all indices $i,j,k,l$ apart into four sets of terms with even exponents, and one set which all have at least one odd exponent (and thus has expectation 0). There are $n$ elements in the first sum, and $n(n-1)$ elements of the second, third, and fourth sum. Thus:
\begin{align*}
\mathbb{E}\Big[\sum_{i=j=k=l} x_i^4\Big] &= n\mathbb{E}\big[\mathcal{X}^4\big]\\
\mathbb{E}\Big[\sum_{i=j,k=l\neq i}x_i^2 x_k^2\Big] &= n(n-1)\mathbb{E}\big[\mathcal{X}^2\big]^2\\
\mathbb{E}\Big[\sum_{\text{remaining}}x_ix_jx_kx_l\Big] &= 0
\end{align*}

and therefore,
\begin{align*}
\mathbb{E}\Big[(\sum_{i=1}^n x_i)^4\Big] &= n\mathbb{E}\big[ \mathcal{X}^4\big] + 3n(n-1)\mathbb{E}\big[\mathcal{X}^2\big]^2\\
&= n\mathbb{E}\big[\mathcal{X}^4\big] + 3n^2 - 3n
\end{align*}

Plugging into the original equation we arrive at a functional form of $\text{Var}[z_n^2]$:

\begin{align}
    \text{Var}\big[z^2_n\big]
    &= \frac{1}{n^3}\mathbb{E}\big[\mathcal{X}^4\big] + \frac{3}{n^2} - \frac{3}{n^3} - \frac{1}{n^2} \nonumber \\
    &= \frac{1}{n^3}\mathbb{E}\big[\mathcal{X}^4\big] + \frac{2}{n^2} - \frac{3}{n^3}
    \label{eq:func-form-var-z-squared}
\end{align}

\subsection{Proof that using the coin-flip distribution yields the lowest-variance estimator of $\frac{1}{n}$}\label{sec:appendix-coin-flip-dist-is-best}
Above, we show that to reduce $\text{Var}\big[z^2_n\big]$ the only knob we can turn is reducing the $4^{th}$ moment, because $\mathbb{E}\big[\mathcal{X}\big] =0$ and $\mathbb{E}\big[\mathcal{X}^2\big]=1$ by construction. The $4^{th}$ moment of the coin-flip distribution $x \sim \mathcal{C} \in \{-1,1\}$ is simply
$$\mathbb{E}[\C^4] = \frac{1}{2} 1^4 + \frac{1}{2} (-1)^4 = 1.$$

Plugging this into Equation~\ref{eq:func-form-var-z-squared} yields, for $x \sim \mathcal{C}$:
$$\text{Var}\big[z^2_n\big] = \frac{2}{n^2} - \frac{2}{n^3}.$$

This holds for all $n\ge1$. Therefore, $\text{Var}\big[z^2_1\big] = 0$, which is easy to confirm. Since variance must always be positive, this means that $\mathbb{E}[x^4] \ge 1$ for all $\mathcal{X}$ satisfying our assumptions, and hence using the coin-flip distribution achieves minimum possible variance in its estimation of $\frac{1}{n}$.

\subsection{Proof that variance scales inversely with vector-length}\label{sec:appendix-more-flips-is-good}
We have another, simpler method for reducing variance: we can increase the number of trials (equivalently, the number of coin flips), and average the results together.

Let $\{z_{ni} | i \in 1,...,d\}$ be a set of $d$ independent draws of $z_n$. Then
$$\mathbb{E}\Big[\frac{1}{d}\sum_{i=1}^{d} z^2_{ni}\Big] = \frac{1}{d}\sum_{i=1}^d \mathbb{E}\big[z^2_{ni}\big] = \mathbb{E}\big[z^2_n\big] = \frac{1}{n}$$

as expected (the average of i.i.d samples is simply the expectation). Similarly:
$$\text{Var}\Big[\frac{1}{d}\sum_{i=1}^{d} z^2_{ni}\Big] = \frac{1}{d^2}\sum_{i=1}^d \text{Var}\big[z^2_{ni}\big] = \frac{1}{d}\text{Var}\big[z^2_n\big]$$
This is a well-known fact of how variance scales with samples.
Therefore, we can decrease the variance of our estimator in two ways: picking a good distribution (coin-flip is best) and also increasing the number of trials.

\newpage
\section{Analysis of Linear Coin Flip Network}\label{sec:linear-analysis}
We now analyze the solution to Equation~\ref{eq:defn-f-star} when our function approximator $f_\phi$ is a linear mapping between states and coin flips.
Our goal is to recover an intuitive understanding of how bonuses are estimated when the model has to generalize across inputs.
For simplicity, we consider the case of a single coin flip per encountered state. We represent each state $\mathbf{s}$ as a $p$-dimensional vector, with $\mathbf{S}$ being the $n \times p$ matrix of all encountered states , and $c \sim \{\text{-}1, 1\}^n$ being the $n$-dimensional vector of sampled coin flips.

Thus, $f_\phi(\mathbf{s}) = \mathbf{s} \cdot \boldsymbol{\phi}$, where $\boldsymbol{\phi}$ is the weight vector that parameterizes $f_\phi$. Under this formulation, Equation~\ref{eq:defn-f-star} reduces to solving the following linear least-squares regression problem:
\[
\boldsymbol{\phi} = \arg\min_{\boldsymbol{\phi'}} \lVert\mathbf{S}\boldsymbol{\phi'} - \mathbf{c}\rVert^2.
\]
For the following derivation, we assume that $\mathbf{S}$ has rank $p$ (and $n > p$) in order to recover a unique solution, however this result can be generalized by replacing the inverse with the pseudo-inverse. The solution to this linear regression \citep{golub1971singular} is
\[
\boldsymbol{\phi}=(\mathbf{S}^T\mathbf{S})^{-1}\mathbf{S}^T\mathbf{c}.
\]
We now rewrite $\mathbf{S}$ using its singular value decomposition \citep{golub1971singular}: $\mathbf{S} = \mathbf{U\Lambda V}^T$, where $\mathbf{U}$ is the $n \times n$ orthonormal ``left singular vector" basis, $\mathbf{V}$ is the $p \times p$ orthonormal ``right singular vector" basis, and $\mathbf{\Lambda}$ is a $n \times p$ rectangular diagonal ``singular value" matrix. Thus, $\mathbf{S^TS} = \mathbf{V\Lambda^T U^TU\Lambda V^T} = \mathbf{V\Lambda^T\Lambda V^T}$. Therefore:
\begin{align*}
\boldsymbol{\phi} &=(\mathbf{S^TS})^{-1}\mathbf{S^T c}\\
&= ( \mathbf{V\Lambda^T\Lambda V^T})^{-1}\mathbf{V\Lambda^T U^T c}\\
&= \mathbf{V}(\mathbf{\Lambda^T\Lambda})^{-1} \mathbf{V^T V\Lambda^T U^T c}\\
&= \mathbf{V}(\mathbf{\Lambda^T\Lambda})^{-1}\mathbf{\Lambda^T U^T c}\\
&= \mathbf{V\Lambda^{-1} U^T c}\\
\end{align*}
where $\mathbf{\Lambda}^{-1}$ is the pseudo-inverse of $\mathbf{\Lambda}$, which in the case of a rectangular diagonal matrix is simply the element-wise reciprocal of the diagonal entries, transposed.


Recall that $f_\phi(s) = \mathbf{s}\cdot \boldsymbol{\phi}$. We can gain more intuition about $f_\phi$ by representing $\mathbf{s}$ using the orthonormal basis $\mathbf{V}$: $\mathbf{s} = \sum_i (\mathbf{s^T\cdot V^T_i}) \mathbf{V^T_i} = \mathbf{pV^T}$ where $\mathbf{p}$ says how much of each basis vector there is in $\mathbf{s}$. Now we can derive a simple formula for the expected inverse-count of $\mathbf{s}$:
\begin{align*}
    \mathbb{E}[\frac{1}{\mathcal{N}(\mathbf{s})}] &= \mathbb{E}[\lVert 2 f_\phi(\mathbf{S})\rVert]^2 \\
    &= \mathbb{E}[\mathbf{s}\boldsymbol{\phi\phi^T}\mathbf{s^T}]\\
    &= \mathbb{E}[\mathbf{pV^TV\Lambda^{-T}U^Tcc^TU\Lambda^{-1}V^TVp^T}]\\
    &= \mathbf{pV^TV\Lambda^{-T}U^T}\;\mathbb{E}[\mathbf{cc^T}]\;\mathbf{U\Lambda^{-1}V^TVp^T}
\end{align*}
Since each element of $\mathbf{c}$ is independent of all others, $\mathbb{E}[\mathbf{cc^T}] = \mathbb{I}$. Furthermore, $\mathbf{U}$ and $\mathbf{V}$ are orthonormal bases and so $\mathbf{U^TU} = \mathbb{I}$ and $\mathbf{V^TV} = \mathbb{I}$. Thus, we get the following simplification:
\[
\mathbb{E}[\frac{1}{\mathcal{N}(\mathbf{s})}] = \mathbf{p}(\mathbf{\Lambda^T\Lambda})^{-1}\mathbf{p^T}.
\]
In other words, when using a linear model, CFN stores in its weights how much of each singular vector is present in the dataset $\mathbf{S}$. When presented with a new state $\mathbf{s}$, it computes an inverse count for each singular vector, and returns a linear combination of those, weighted by how much of that singular vector is present in $\mathbf{s}$.

This matches intuition in the tabular case. If each $s \in S$ is represented by a one-hot vector, then $\mathbf{S^TS}$ is a diagonal matrix that counts how many times each unique state has been seen, and $\mathbf{p}(\mathbf{\Lambda^T\Lambda})^{-1}\mathbf{p^T}$ returns the the inverse count of a given state. In the more general linear case, counting simply happens over a different basis than the identity matrix.
\newpage
\section{Environment details}\label{sec:appendix-environment-details}
\paragraph{Visual Gridworld.} We used a $42$x$42$ gridworld, observations were $84$x$84$ images (each grid-cell was visualized using $2$x$2$ pixels). The player always started the episode at the bottom-left and the goal was at the top-right. Episodes lasted a maximum of $150$ steps, unless the agent reached the goal, in which case the episode terminates with a sparse reward of $1$. For the stochasticity experiments in Figure~\ref{fig:gridworld-learning-curves}(right), the maximum number of steps per episode was determined by $\frac{150}{1-\eta}$ where $\eta$ is the action-noise probability. We did this to make sure that more stochastic versions of the task still had long-enough episodes to be solvable by a reasonably good agent: with this scaling the agent has the same number of actions under its own control in a given episode, independent of $\eta$.
\paragraph{Visual Taxi.} We used two variants of the \textsc{Taxi}---one with a $5$x$5$ grid (as in the default version) and another with a $10$x$10$ grid \citep{diuk2008object}; episodes lasted a maximum of $50$ steps in the former and $100$ in the latter case. Similar to visual gridworld, the agent observed images of the game state; to show that the passenger was inside the taxi, we shaded the taxi differently and included a black border around the image.
\paragraph{Fetch and Adroit Manipulation.} As mentioned in the main paper, we first sparsified the reward function---this means that there are no shaping rewards for reaching the object or for moving it to non-goal locations. To set the goal locations for the non-default versions of the tasks, we first visualized the environment and rendered the effect of random actions. The exact goal locations that we settled on can be found in our linked code (file \texttt{wrappers.py}). In the \textsc{Relocate} task, we truncate the episode when the ball leaves the table.
\paragraph{Ant U-Maze.} D4RL randomly sets the goal location to a small distribution around $(x=0, y=8)$ and the ant's location to a small distribution around $(x=0,y=0)$. Predictably, we used the sparse-reward version of the task. Episodes last a maximum of $1000$ steps.
\paragraph{\textsc{Montezuma's Revenge}} We followed the experimental protocol of \citet{machado2015domain} which means that we used sticky actions, a frame stack of $4$, action repeat of $4$, grayscale images of shape $84$x$84$ and a training budget of $200$ million frames ($50$ million agent-environment interactions).

\newpage
\section{Hyperparameters, Architecture and Training Details}\label{sec:appendix-hyperparameters}

The final values of hyperparameters for all experiments have been set as the default values in our configuration files, which can be found in the \texttt{configs/*.gin} files in our codebase. The file \texttt{intrinsic\_motivation/intrinsic\_rewards.py} contains architectural details such as number of layers and layer sizes, and unless otherwise noted are the same as presented in \citep{Taiga2020On}. The neural network architecture of CFN is chosen to match that of RND's prediction network.

All hyperparameters \textit{not} listed are chosen to match the default Rainbow \cite{hessel2018rainbow} and SAC \cite{haarnoja2018soft} implementations. For each task group, the tested hyperparameters are listed; multiple value indicate a grid search, with the chosen value listed in bold. On Montezuma's Revenge we use the best reported RND and PixelCNN hyperparameters from \citep{Taiga2020On}.

\subsection{Shared Hyperparameters}
We used Rainbow for Visual Gridworld (Section~\ref{sec:experiements_gridworld_rl}) and Montezuma's Revenge (Section~\ref{sec:monte_results}) and SAC for all the continuous control experiments (Section~\ref{sec:experiments_cc}). The hyperparameters for these base agents are reported below:

\begin{center}
\begin{tabular}{ c|c|c } 
 \hline
 Rainbow Hyperparameter & Gridworld & Atari \\ 
 \hline
 Discount Factor $\gamma$ & 0.99 & 0.99 \\
 Adam LR & \textbf{1.25e-4}, 1e-5 & \textbf{1.25e-4}, 6.25e-5 \\
 Adam $\epsilon$ & 1.5e-4 & 1.5e-4 \\
 Multi-step return $n$ & 3 & 3 \\
 Min history to start learning & 1,000 & 20,000\\
Distributional Atoms & 51 & 51\\
Distributional Min/Max values & $\pm10$ & $\pm10$\\
Batch Size & 32 & 32\\
\end{tabular}
\end{center}

\begin{center}
\vspace{0.1cm}
\begin{tabular}{ c|c|c|c } 
 \hline
 SAC Hyperparameter & Fetch & Adroit & Ant \\ 
 \hline
 Discount Factor $\gamma$ & 0.99 & 0.99 & 0.99 \\
 Adam LR & 3e-4 & \textbf{1e-4} 3e-4 & 3e-4 \\
 Adam $\epsilon$ & 1e-8 & 1e-8 & 1e-8 \\
 Multi-step return $n$ & 3 & 3 & 3 \\
 Min history to start learning & 10,000 & 10,000 & 10,000\\
 SAC Reward Scale Factor & 1.0 & 1.0, 3.0, 10.0, \textbf{30.0} & 0.1\\
 Batch Size & 256 & 256 & 256
\end{tabular}
\end{center}

\subsection{CFN Hyperparameters}
\label{sec:cfn-hypers}

\begin{center}
\begin{tabular}{ c|c|c|c|c|c } 
 \hline
 Hyperparameter & Gridworld & Fetch & Adroit & Ant & Atari \\ 
 \hline
Intrinsic Reward Scale & \shortstack{0.001, 0.003 \\ \textbf{0.01}, 0.03} & \shortstack{\textbf{0.001}, 0.003 \\ 0.01, 0.03} & \shortstack{\textbf{0.001}, 0.003 \\ 0.01, 0.03} & \shortstack{0.001, 0.003 \\ \textbf{0.01}, 0.03} & \shortstack{0.001, 0.003 \\ \textbf{0.01}, 0.03} \\
\hline
Reward Normalization$^*$ & \textbf{Yes}, No & No & No & No & No \\
\hline
CFN learning rate & 1e-4 & 1e-4 & 1e-4 & 1e-3 & 1e-4 \textbf{1e-5} \\
\hline
CFN replay buffer size & 1e6 & 1e6 & 1e6 & 1e6 & 2e7\\
\hline
CFN batch size & 1024 & 1024 & 1024 & 1024 & 512\\
\hline
CFN update period & 1 & 1 & 1 & 1 & 4\\
\hline
Number of coin flips $d$ & 20 & 20 & 20 & 20 & 20\\
\hline
\end{tabular}
\end{center}

\paragraph{$^*$ Reward normalization.} In Visual Gridworld, we found it helpful to use reward normalization \citep{Burda2018}, which normalizes the exploration bonus by subtracting its running mean and dividing by its running variance.

\subsection{CFN replay buffer size}\label{sec:appendix-cfn-replay-size}
Note that we used a much larger CFN replay buffer for Montezuma's Revenge. Usually, a small FIFO queue is used to implement the replay buffer. As shown by \citet{zhang2017deeper}, larger replay buffers hurt RL performance because it makes the Q-learning updates more off-policy. Since the CFN objective is a standard regression problem (and does not use bootstrap targets), larger buffers almost always improve performance. Furthermore, shorter CFN buffers often cause high-count states to be removed from the replay buffer and eventually re-appear as novel to the agent; a problem that is mitigated with larger buffer sizes. In future work, we would like to revert to smaller replay buffers by implementing a ``forgetting'' strategy in which low novelty states are discarded from replay with higher probability.

\begin{figure}
    \centering
    \includegraphics[width=0.7\linewidth]{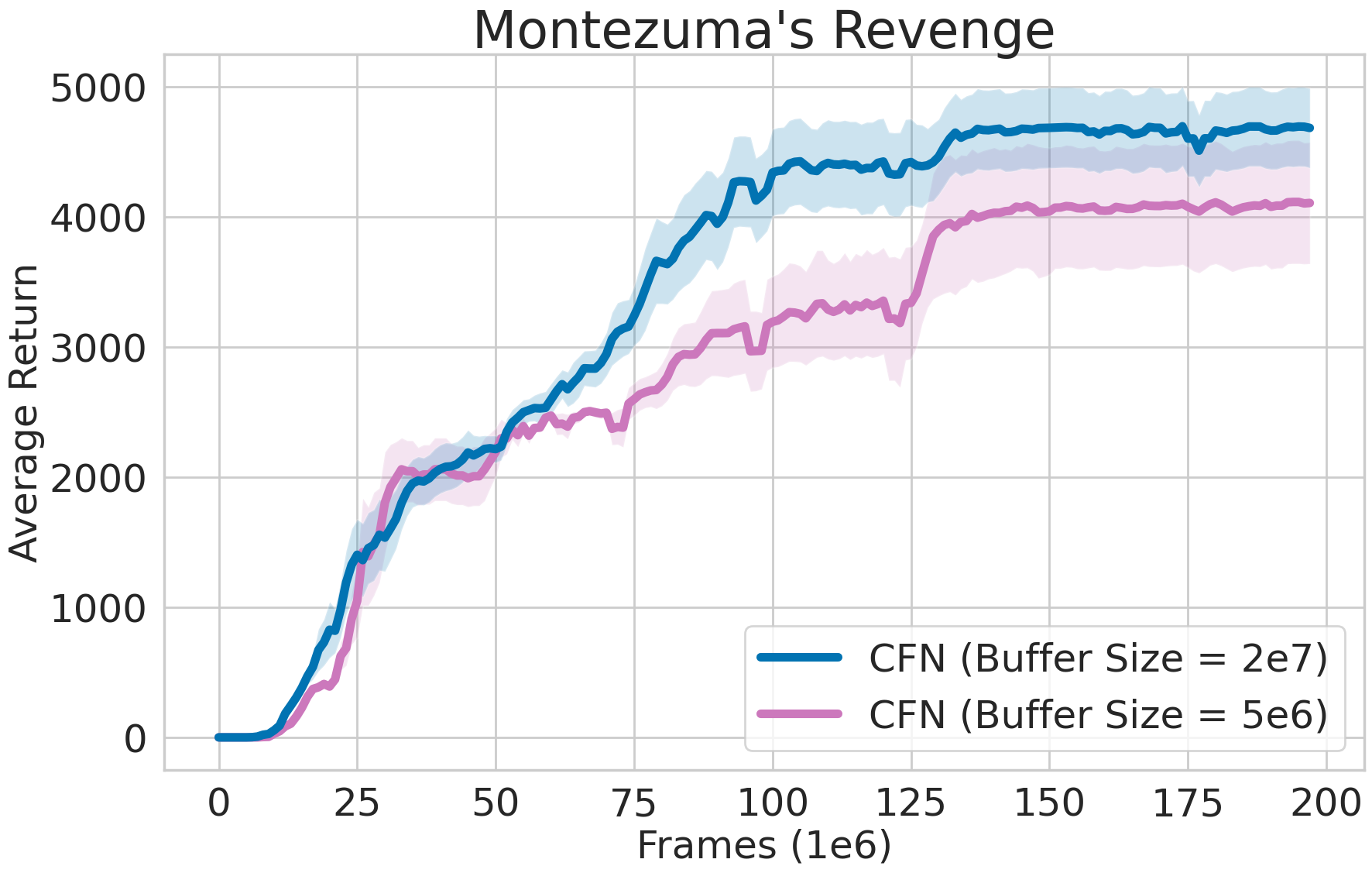}
    \caption{Analyzing the impact of large replay buffer for CFN: while a very large buffer size of $2e7$ leads to better performance, a smaller buffer size of $5e6$ still performs well. Results are averaged over $6$ random seeds.}
    \label{fig:monte_big_vs_medium_mem}
\end{figure}

Figure~\ref{fig:monte_big_vs_medium_mem} shows that while a very large replay buffer does indeed yield better performance in Montezuma's Revenge, a moderately sized replay buffer also performs respectably.

\subsection{PixelCNN Hyperparameters}

\begin{center}
\begin{tabular}{ c|c|c } 
 \hline
 Hyperparameter & Gridworld & Atari \\ 
 \hline
Intrinsic Reward Scale & 0.1, \textbf{0.5}, 1.0 & 0.1\\
Prediction Gain Scale & 0.1, \textbf{0.5}, 1.0, 5.0 & 1.0 \\
PixelCNN learning rate & 1e-4 & 1e-4
\end{tabular}
\end{center}

\subsection{RND Hyperparameters}
\begin{center}
\begin{tabular}{ c|c|c|c|c|c } 
 \hline
 Hyperparameter & Gridworld & Fetch & Adroit & Ant & Atari \\ 
 \hline
Intrinsic Reward Scale & \shortstack{5e-5,\textbf{1e-4},\\5e-4,1e-3} & \shortstack{\textbf{5e-5},1e-4,\\5e-4,1e-3} & \shortstack{\textbf{5e-5},1e-4,\\5e-4,1e-3} & \shortstack{\textbf{5e-5},1e-4,\\5e-4,1e-3} & 5e-5 \\
\hline
RND learning rate & 1e-4 & 1e-4 & 1e-4 & 1e-4 & 1e-4 \\
\hline
\end{tabular}
\end{center}

\subsection{Compute Resources}
The wallclock time of all intrinsic reward methods is roughly similar, however on Atari domains CFN requires significantly more memory---see Section~\ref{sec:cfn-hypers} for discussion. All experiments are performed on a \textsc{slurm} cluster using nodes equipped with $4$ CPUs and $1$ $3090$-Ti NVIDIA GPU. We reserve $16$GB of memory for all experiments except for CFN's Montezuma's Revenge, where we reserve $160$GB.

\newpage
\section{Additional Experiments} \label{sec:appendix-additional-experiments}
\subsection{Counts on Visual Taxi}\label{sec:appendix-taxi-counts}
We report count reconstruction-accuracy on visual versions of both the 5x5 and the 10x10 \textsc{Taxi} environments in Figure~\ref{fig:taxi-random-policy-counts}. See Appendix~\ref{sec:appendix-environment-details} for environment details. These tasks are visually more complex and have many more possible states than Visual Gridworld. Since the policy used to collect data is a confounding variable when comparing counting accuracy, in these experiments all interaction is performed with a random policy, and only the bonus modules are trained. For both domains, we present bonuses computed after $200,000$ interactions, with each method taking one training step per interaction. In this time, the random policy visits approximately $900$ unique states in the 5x5 domain, and approximately $6,500$ unique states in the 10x10 domain. We note a similar trend as in Gridworld, where CFN procudes more accurate bonuses than PixelCNN and RND.

\begin{figure}[h]
\centering
\vspace{1cm}
\includegraphics[width=\linewidth]{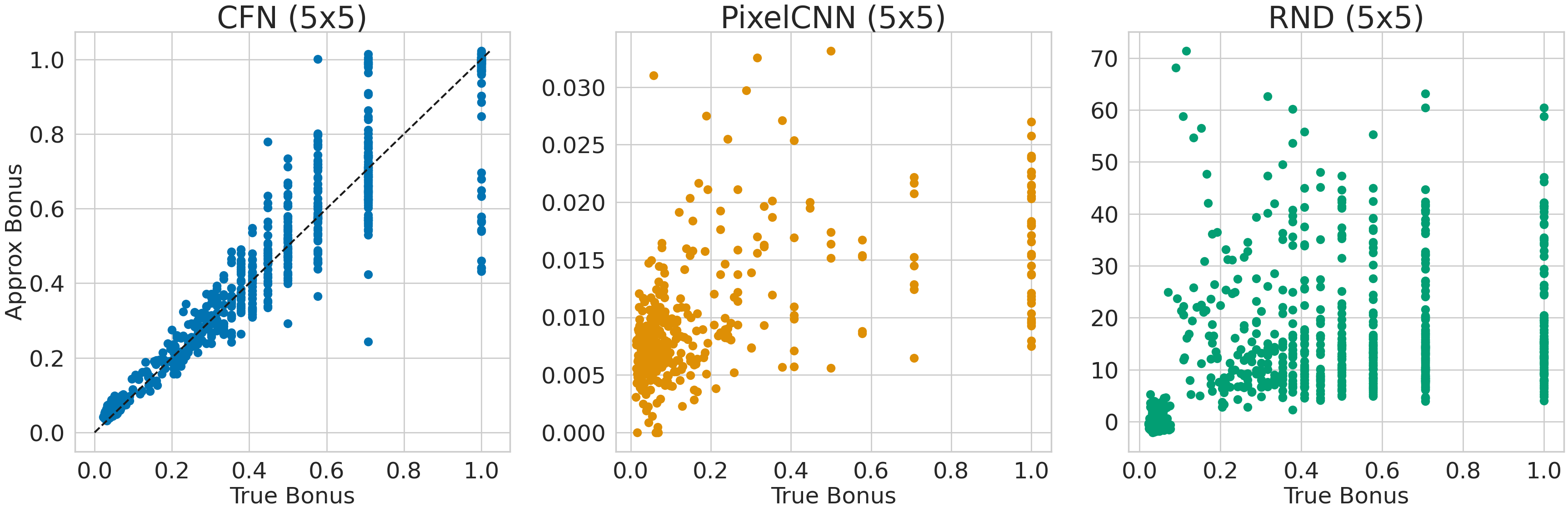}\newline
\vspace{0.5cm}
\includegraphics[width=\linewidth]{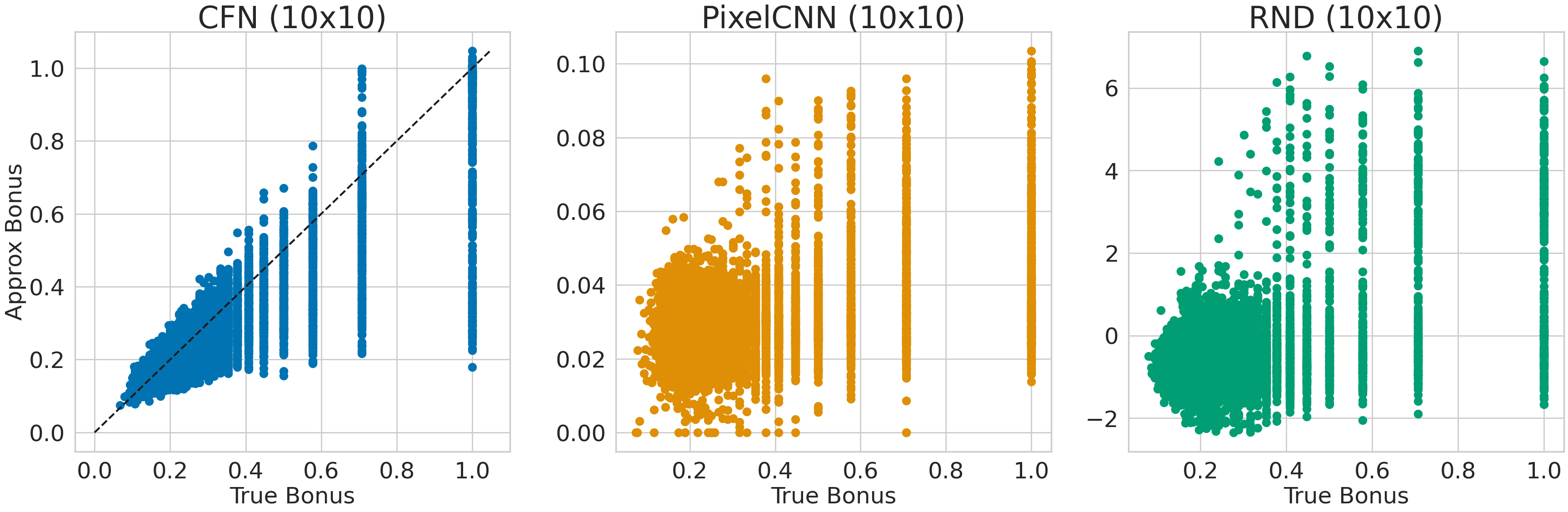}
\vspace{-0.5cm}
\caption{Bonuses for CFN, PixelCNN, and RND on both Taxi domains.}
\label{fig:taxi-random-policy-counts}
\end{figure}

\newpage
\subsection{Effect of Random Prior without Coin Flips}\label{sec:app-zero-flips}

Can the accuracy of the predicted exploration bonuses in Figure~\ref{fig:gridworld-counts} (main paper) solely be attributed to the use of the random prior (discussed in Section~\ref{sec:cfn_randomprior})? To answer this question, we devised an experiment in which we removed the randomness introduced by the coin-flip vectors by setting them to $\{0\}^d$ instead of sampling them from $\{-1,1\}^d$.

\begin{figure}[h]
\centering
\vspace{1cm}
\includegraphics[width=0.5\linewidth]{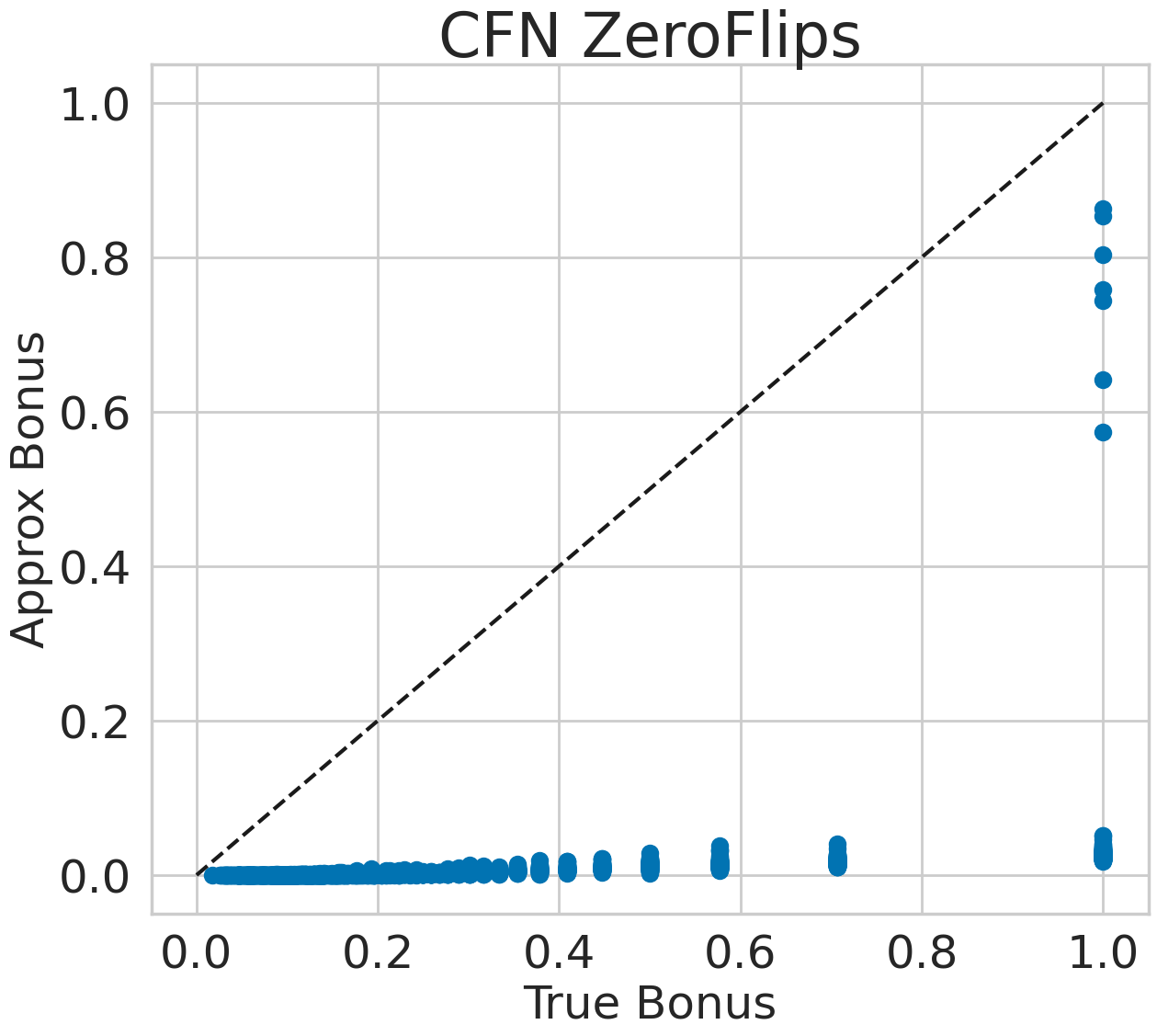}
\caption{CFN bonus prediction versus ground-truth bonus on the $42$x$42$ Visual Gridworld domain using $c \sim \{0\}^d$.}
\label{fig:zero-flip-counts}
\end{figure}

Figure~\ref{fig:zero-flip-counts} shows the result of this experiment: states seen for the first time get a high exploration bonus, implying that the random prior initializes their pseudocount near $1$ (as intended). On the other hand, states visited more than once get almost no exploration bonus; this highlights the importance of the full CFN objective to get an exploration bonus that falls off smoothly as $1/\sqrt{\N(s)}$.

Furthermore, notice that states with true bonus of $1$ form two distinct clusters---with high and low predicted bonus respectively. We posit that states in the lower bonus cluster were encountered less recently and thus have been sampled by CFN, which has wiped out the effect of their optimistic initialization.


\newpage
\subsection{Prioritization and Random Prior ablation count plots}\label{sec:appendix-per-rp-ablation}
\begin{figure}[h]
\centering
\vspace{1cm}
\includegraphics[width=0.9\linewidth]{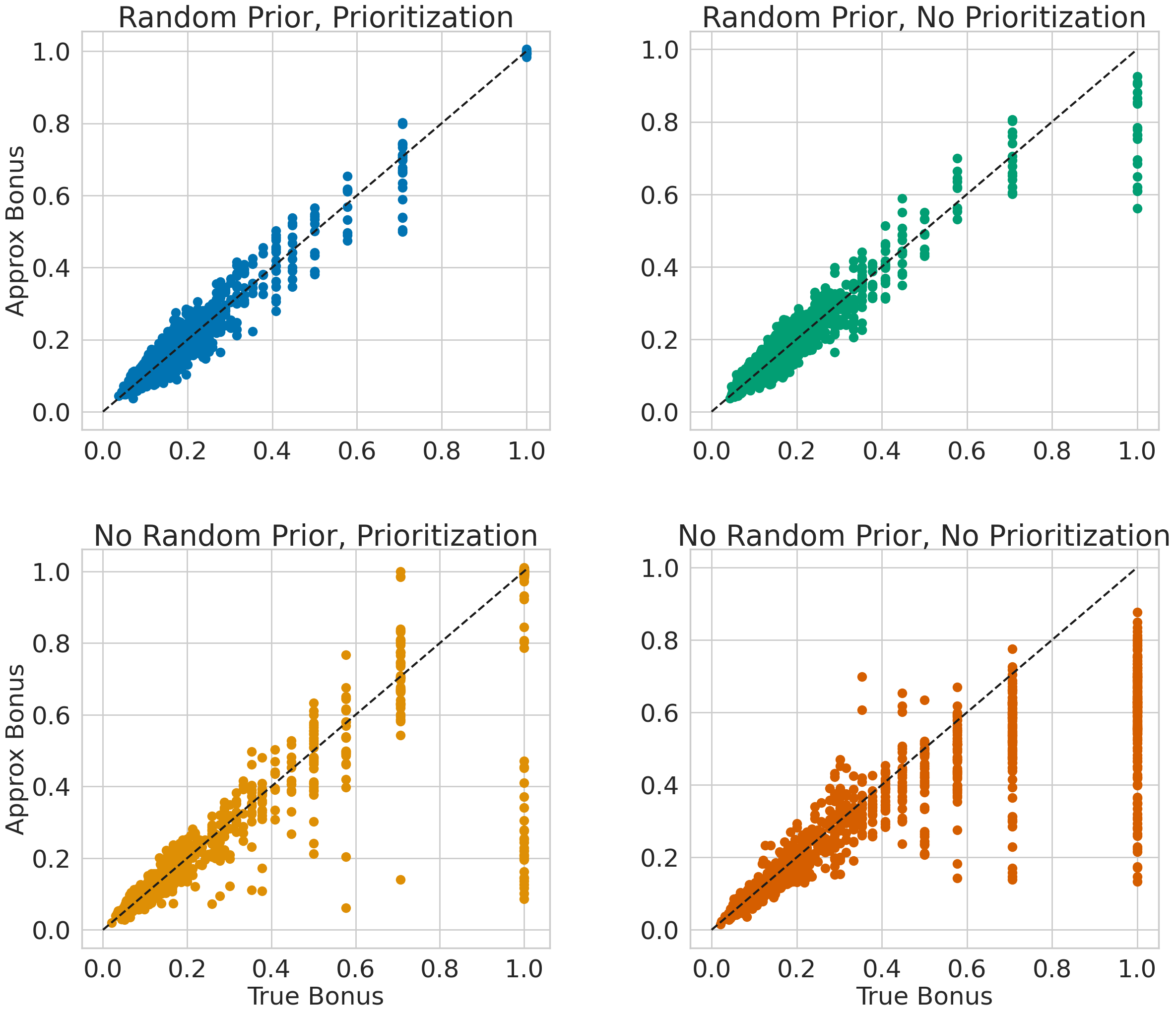}
\caption{True versus approximate bonus of including/excluding random prior and prioritization.}
\label{fig:gridworld-ablation-4-count-plots}
\end{figure}

Figure~\ref{fig:gridworld-ablation-4-count-plots} shows the importance of prioritized sampling and optimistic bonus initialization. Using both methods results in the most accurate bonus predictions across the entire range of novelty. When only random prior is used, CFN does not train as frequently on novel states, and thus underestimates their count. When only prioritization is used, a novel state may be first inserted into the dataset with low novelty (and thus low priority); so, the state may not be sampled for training, which retains its underestimate of novelty. 
When neither prioritization nor random prior are used, bonuses are still accurate for states observed more than $5$ times, but are inaccurate for very novel states.

\end{document}